%% file: paper.tex
\documentclass[10pt]{handshaketemplate_v2}
\usepackage[authoryear,round]{natbib}
\usepackage{float}
\usepackage{longtable}
\usepackage{needspace}
\usepackage{fvextra}
\newif\ificlr 
\let\originalsection\section
\renewcommand{\section}{\Needspace{9\baselineskip}\originalsection}
\graphicspath{{figures/}}
\title{StudentBench: AI and human tutoring yield equivalent GRE learning gains}
\shorttitle{StudentBench: AI and human tutoring yield equivalent GRE learning gains}
\author{Curtis Northcutt, Inaara Hasmani, Kevin Feng, Trevor Khangi, Andreas Plesner, Jonas Mueller}
\paperdate{September 2026}
\handshakelogo{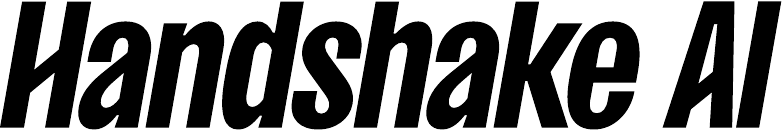}
\pdfpapertitle{StudentBench: AI and human tutoring yield equivalent GRE learning gains}
\pdfpaperauthor{Curtis Northcutt, Inaara Hasmani, Kevin Feng, Trevor Khangi, Andreas Plesner, Jonas Mueller}

\DeclareRobustCommand{\figpanelref}[2]{\mbox{\hyperref[#1]{\ref*{#1}#2}}}
\newcommand{\doi}[1]{doi: \href{https://doi.org/#1}{{\urlstyle{rm}\nolinkurl{#1}}}}
\newcommand{\paperfigure}[4]{\begin{figure}[t]\centering\IfFileExists{figures/#1.pdf}{\includegraphics[width=\linewidth,height=#4\textheight,keepaspectratio]{#1.pdf}}{\fbox{\parbox[c][0.12\textheight][c]{0.94\linewidth}{\centering Figure in preparation: #1}}}\caption{#2}\label{#3}\end{figure}}
\begin{abstract}
Artificial intelligence offers an unprecedented opportunity to augment human capabilities, yet progress at the frontier has focused primarily on advancing model capabilities. We introduce StudentBench, a suite of AI teaching evaluations and a public platform that enables large-scale data collection with over 175,000 student--AI messages to study whether large language models (LLMs) produce learning gains equivalent to human tutoring. Using StudentBench, we measured learning gains on Quantitative and Verbal GRE questions across 2,383 human participants receiving AI tutoring, human tutoring, or no tutoring. We establish that AI tutoring is statistically equivalent to expert human tutoring for GRE learning gains ($p{=}.015$), and in five of the seven GRE domains, the best performing AI tutor surpassed the human tutor, on average. In a second study, expert human tutors compared LLM-generated lesson plans and practice problems through 2,028 pairwise rubric evaluations. Together, the two studies clearly separate AI tutors across: (1) lesson planning, (2) practice-problem creation, (3) conversational pedagogy, (4) cost, and (5) engagement. Surprisingly, one AI tutor achieved learning gains equivalent to human tutoring ($p{=}.044$) at $918\times$ lower cost (\$0.0052 for AI versus \$4.81 for human, per percentage point gained). For Quantitative GRE sessions, faster AI replies correlated with more student messages, more messages with more correct practice, and more correct practice with larger learning gains (all $p{\le}.002$). To support future research, we open-source the de-identified data collected in our studies.
\end{abstract}
\begin{document}
\maketitle
\input{main}
\label{maintextend}
\label{main_end}
\input{appendix_ethics}
\input{appendix_reproducibility}
\bibliographystyle{plainnat}
\bibliography{references}
\clearpage
\appendix
\counterwithin{figure}{section}
\counterwithin{table}{section}
\section*{Appendix}
\suppressfloats[t]
\input{appendix_grouped}
\end{document}

%% file: main.tex
\begin{figure}[!t]\centering
\includegraphics[width=\linewidth,height=0.76\textheight,keepaspectratio]{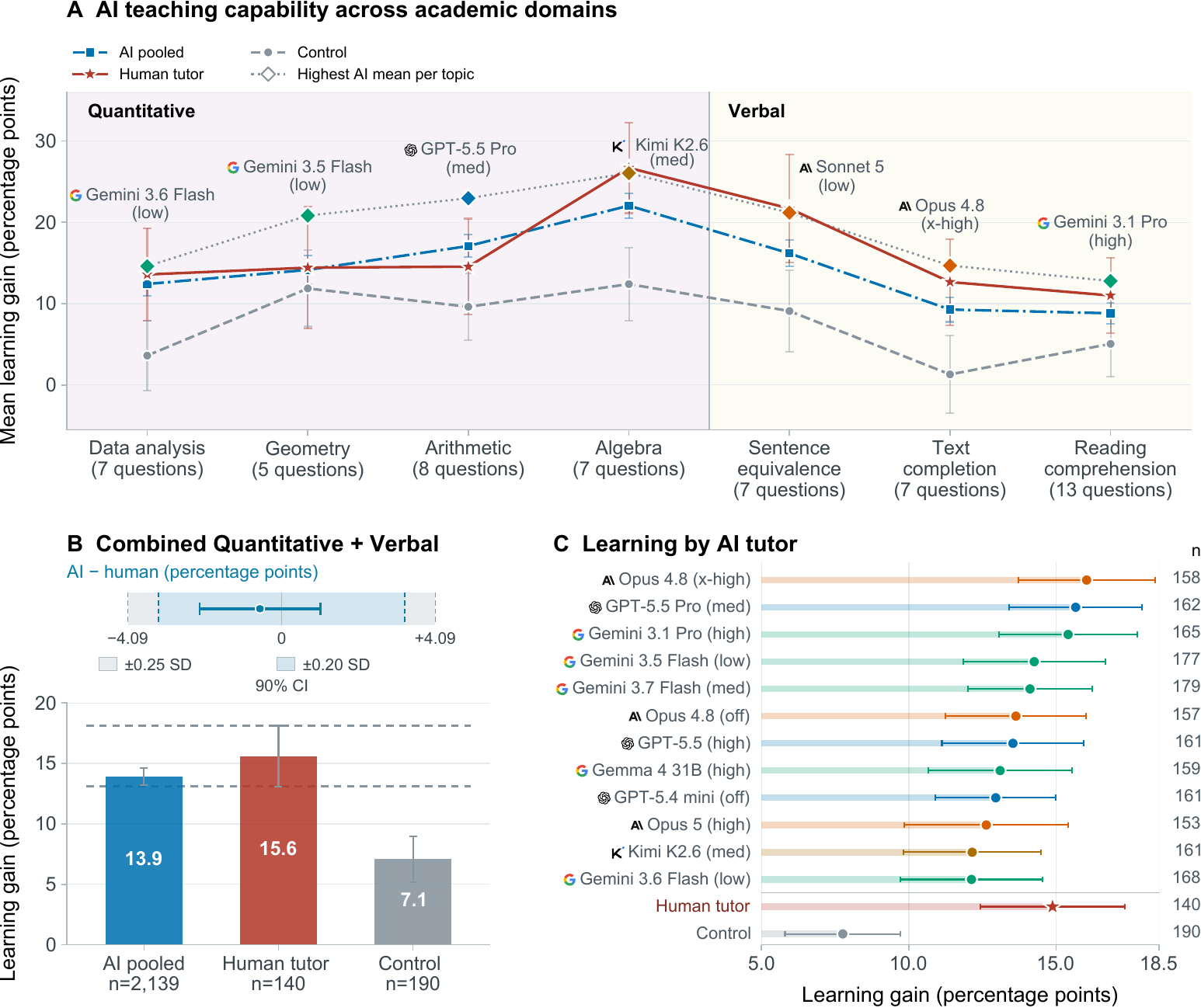}
\caption{\textbf{Learning across GRE domains and tutors.} \textbf{A,} Mean pre-to-post test score change within each GRE domain under different tutoring conditions:  human tutoring, AI tutoring (results pooled across all AI tutors), or no tutoring control. Error bars depict 95\% Student-$t$ confidence intervals. Diamonds mark the top performing model in each domain; the dotted line traces the best AI performance across domains. Question counts refer to each assessment (pre-test, post-test). \textbf{B,} Learning gains (post $-$ pre test scores) pooled across Quantitative and Verbal, with 95\% intervals; $n$ denotes the number of sessions under each tutoring condition. Dashed guides indicate the confidence interval for the gain from human tutoring. The inset shows the adjusted AI $-$ human difference and 90\% confidence interval within $\pm0.25$ pooled-SD equivalence bounds ($p{=}.015$); the interval also lies within the stricter $\pm0.20$-SD bounds ($p{=}.023$). \textbf{C,} Learning gains for 12 AI tutors, adjusted for pre-test score and section, with 95\% HC3 intervals. See Appendix~\ref{app:rankings} for learning gains (Table~\ref{tab:combined_learning_complete}) and AI tutor leaderboards for all seven domains (Section~\ref{app:category_leaderboards}).}\label{fig:learning}
\end{figure}

\input{introduction}

\section{Measuring learning and teaching}\label{sec:measurement}\label{sec:prompt_design}\label{sec:study_design}\label{sec:scores}

\begin{figure}[t]
    \centering
    \includegraphics[width=1\linewidth]{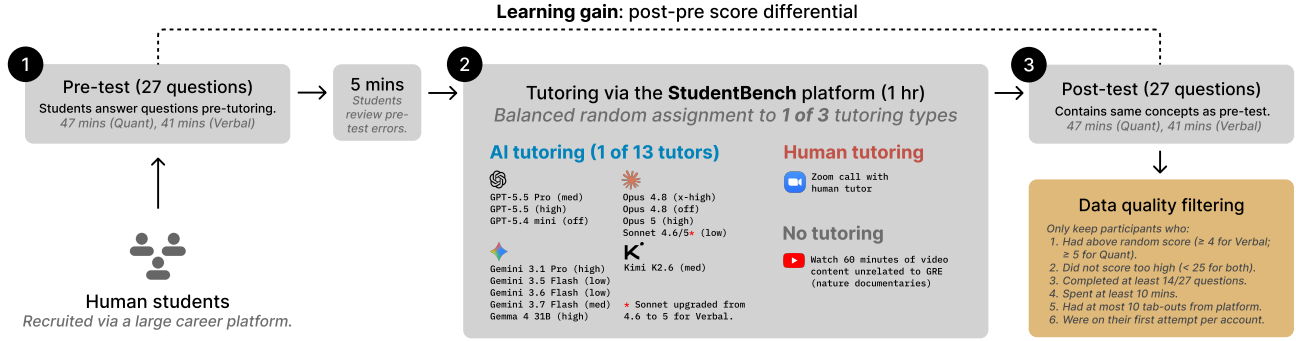}
    \caption{\textbf{Experimental design of the learning gain study.} \textbf{1,} Students completed either a 27-question Quantitative or Verbal \textbf{pre}-test, with eligibility and effort screening, and then spent five minutes reviewing incorrect answers. \textbf{2,} Students were assigned to one of three tutoring conditions: AI tutoring (with one of 13 LLM-based tutors; Table~\ref{tab:models}), human tutoring, or no tutoring control. \textbf{3,} After tutoring, students completed a \textbf{post}-test with the same length and concepts as the pre-test.}
    \label{fig:placeholder}
\end{figure}

To evaluate how effectively AI tutors teach, we compared student learning gains across three conditions: AI tutoring, human tutoring, and no tutoring control.
In each session, students took a pre-test, spent one hour in their assigned condition, and then took a post-test with different questions (Figure \ref{fig:placeholder}). Each assessment matched the length and timing of the GRE exam: 27 questions in 47 minutes for Quantitative and 41 minutes for Verbal. 
After applying the data quality filters shown in Figure \ref{fig:placeholder}, our analysis included 2,469 Quantitative and Verbal sessions from 2,383 students. Appendix~\ref{app:interfaces} shows the study interfaces in which students received AI or human tutoring and took assessments, and how experts compared lesson plans (Figures~\ref{fig:interface_tutoring}--\ref{fig:interface_lesson_comparison}).

The pre-tests and post-tests for Quantitative and Verbal were crafted by former GRE exam creators from ETS and Kaplan, who designed new questions to match the difficulty, coverage, order, categories and question types of prior GRE exams. Old GRE questions were never used because LLMs may have been trained on publicly available former GRE questions~\citep{brown2020language}.

AI tutoring encompasses lesson planning, text-based conversational tutoring, and practice problem generation. Students were randomly assigned a tutoring condition and those assigned an AI model were not told which. Students assigned to the control condition watched educational videos unrelated to the GRE. Human tutors taught one-to-one in live video calls. Both human and AI tutors were given the student's pre-test mistakes to guide lesson planning and teaching, but were not given access to the questions on the post-test. Across all conditions, students spent about five minutes reviewing their mistakes on the graded pre-test. 

Every AI tutor is built with two prompts for (1) lesson planning and problem creation, and (2) interactive tutoring live with the student. We use minimal software scaffolding and prompt adjustments (Appendix~\ref{app:prompts}, Fig.~\ref{fig:prompt_comparison}), so we can measure LLM teaching capabilities with limited influence from our design. The AI tutors determined what to teach, planned lessons, tutored students, and generated practice problems and answer keys. 

Our results establish what AI tutors can achieve with low guidance; further prompt adjustment and software scaffolding are likely to improve AI tutor performance. For example, on ARC-AGI-3, changes in harnesses and software scaffolding led to substantial performance gains \citep{karten2026primeagent}.

The study took place over a two-month period in summer 2026. Through Handshake’s platform, we invited approximately 70,000 randomly selected students across all majors, primarily aged 18 to 23. Within the AI condition, StudentBench randomly assigned students to different \emph{AI tutors}, keeping counts balanced across available AI tutors. Each AI tutor is defined by its model and reasoning setting. New AI tutors entered the study as they became available.

We define \emph{learning gain} as post-test score minus pre-test score, measured in percentage points, where score is the percentage of questions the student answered correctly.  Unanswered questions count as incorrect, as in the GRE. To account for differences in average pre-test score across conditions when comparing learning gains, we use the analysis of covariance (ANCOVA) approach \citep{vickers2001}. See Appendix~\ref{app:learning_analysis} for details. For AI $-$ human equivalence comparisons, we also adjust for assessment form and section and account for shared tutors and repeated students using CR2 covariance with Satterthwaite degrees of freedom \citep{pustejovsky2018}. We test equivalence using standard two one-sided tests at $\alpha{=}0.05$ \citep{schuirmann1987,lakens2017}. We use equivalence bounds of $\pm0.25$ pooled standard deviations of learning gains, consistent with standardized bounds used in prior analyses of reading comprehension and treatment outcomes \citep{schwabe2022equivalence,steinert2017equivalence}. Equivalence requires the entire 90\% confidence interval for the AI $-$ human difference to fall within these bounds, corresponding to $\pm4.09$ percentage points for the combined GRE comparison. To evaluate human equivalence for individual (non-pooled) AI tutors, we use the same model and bounds in separate equivalence tests at $\alpha{=}.05$, without correction for multiple comparisons across tutors.

In a second study, 51 expert human tutors completed 2,028 pairwise reviews of AI-generated lesson plans and their embedded practice problems based on 381 student pre-tests. Each comparison paired one study plan with another generated by a different AI tutor from the same pre-test. For each pair, reviewers answered eight comparison questions: five about lesson planning, used for Figure~\figpanelref{fig:rubrics}{A}, and three about practice problems, used for Figure~\figpanelref{fig:rubrics}{B}. Appendix~\ref{app:rubric} describes the selected pre-tests, generation conditions, and individual criteria.

\paragraph{Data quality and assessment integrity.}
Sessions were excluded for incompleteness, low effort, rapid responses, below-random performance, frequent tab switching, and pre-test scores too high for improvement ($>24/27$). See Appendix~\ref{app:data_quality} for data filter details and Table~\ref{tab:data_quality} for exclusion counts.
To avoid reward hacking in AI tutors~\citep{amodei2016concrete}, AI tutors were never given access to the post-test at any point and the prompts explicitly prohibited speculating about post-test content (Appendix~\ref{app:prompts}). 

\section{AI and human tutoring yield significant and comparable score gains}\label{sec:learning}
Human tutoring and AI tutoring pooled across all AI tutors produced equivalent mean learning gains across the combined Quantitative and Verbal sessions ($p{=}.015$, $\pm0.25$-SD bounds; also significant under the tighter $\pm0.20$-SD bounds, $p{=}.023$; 1.59 d.f.). The combined AI $-$ human difference was $-0.58$ percentage points (90\% CI $[-2.18,1.03]$). At the $\pm0.25$-SD margin, equivalence was also established for Quantitative, but not Verbal.

Our pooled AI result spans 13 capability-varied AI tutors, including non-frontier and open-weight models. After adjustment for pre-test score, the AI $-$ control difference in learning gain was 6.86 percentage points in Quantitative (95\% CI $[4.02,9.69]$) and 5.47 in Verbal ($[2.46,8.47]$). Gains with human tutoring were also higher than those in the control condition in both sections. Across both sections, the AI $-$ control difference was 6.15 percentage points ($[4.08,8.21]$). That is about 1.5--2 more correct answers (out of 27) than in the control.

Figure~\figpanelref{fig:learning}{A} shows the frontier of AI tutoring across seven GRE domains. The red line shows average human tutor performance in each domain, the blue line shows pooled AI tutor performance, and the dotted line connects the highest AI mean learning gain in each domain, with a different tutor leading each domain. The gray line shows the control condition, and error bars show 95\% confidence intervals for these three conditions.

This visualization helps us see where the strongest AI tutors stand relative to humans. In Quantitative, pooled AI and human mean learning gains are generally close, while the best AI tutors have higher mean gains than humans in three of the four domains. In Verbal, human mean gains are close to those of the best AI tutors, but remain above the pooled AI mean in all three domains. This contrast helps us understand where AI tutoring capabilities are strongest and where human tutors still have an advantage.

We were curious which AI tutors teach high-performing students best. To study this, we used item response theory (IRT) to estimate proficiency on the Quantitative pre-test \citep{lord1980irt} and compared learning gains among students in the top 25\%. These students have less room for improvement, yet four of the five highest mean gains came from Gemini tutors, led by Gemini 3.5 Flash. This exploratory finding may suggest a differentiated strength of the Gemini family of models in teaching high performers.

See Appendix~\ref{app:protocol} for tutor details; Appendix~\ref{app:statistics} for sensitivity checks; Appendix~\ref{app:rankings} for learning-gain estimates, domain rankings, and starting-proficiency analyses; and Appendix~\ref{app:section_frontiers} for learning gains by section.

\section{Three evaluations of AI teaching}\label{sec:teaching}
\begin{figure}[!t]\centering
\includegraphics[width=\linewidth,height=0.73\textheight,keepaspectratio]{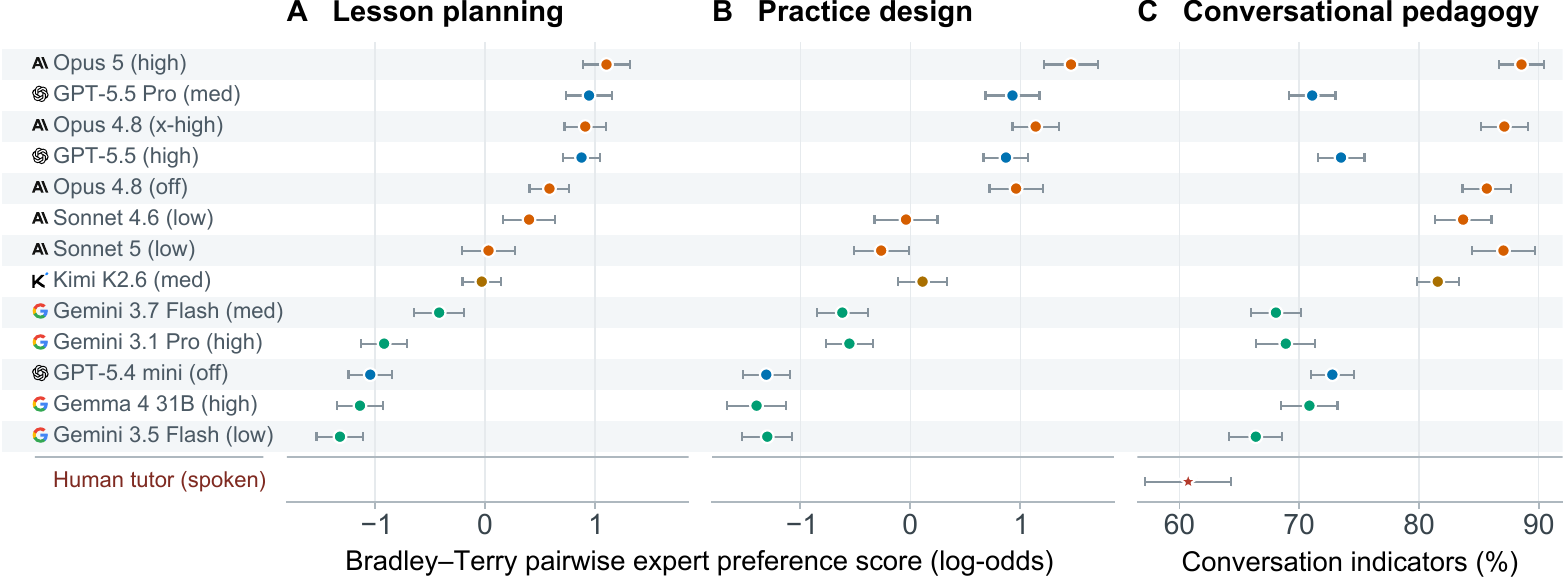}
\caption{\textbf{Three evaluations of AI teaching.} Rows are ordered by lesson-planning score. \textbf{A, B,} Expert preferences for lesson planning and practice-problem creation, fitted with separate centered Bradley--Terry models; score differences are log-odds. Ratings marked as ties are excluded. Planning uses 9,577 ratings across 381 pre-tests; practice design uses 5,265 across 380. Whiskers are pointwise 95\% intervals clustered by pre-test. \textbf{C,} Mean prevalence of six fixed conversation indicators across 1,971 AI and 135 human transcripts, with 95\% Student-$t$ intervals. The human reference is spoken tutoring; AI sessions use typed chat. Scores should be compared only within panels; blank cells mean not evaluated. Colors identify model families. See Appendices~\ref{app:rubric} and~\ref{app:conversation} for evaluation details.}\label{fig:rubrics}
\end{figure}

In this section, we present three evaluations of AI teaching. For each pair of AI lesson plans, expert reviewers answered eight comparison questions: five about lesson planning, combined to create the leaderboard in Figure~\figpanelref{fig:rubrics}{A}, and three about practice problems, combined to create the leaderboard in Figure~\figpanelref{fig:rubrics}{B}. We separate the two groups because lesson planning and practice-problem creation are distinct teaching capabilities. We also evaluate pedagogical characteristics of tutoring conversations to create the leaderboard in Figure~\figpanelref{fig:rubrics}{C}.

\subsection{Lesson planning and practice-problem creation}\label{sec:planning_practice}

The \textit{Lesson planning leaderboard} evaluates concept relevance, grouping, prioritization, time allocation, and test-taking strategies \citep{anderson1995,corbett1995}. The \textit{practice-problem design leaderboard} evaluates alignment with the student's mistakes, appropriate difficulty, and accuracy of examples and answer keys. Appendix~\ref{app:rubric} gives the criteria (Table~\ref{tab:rubric}), and detailed rankings (Figure~\ref{fig:criterion_bt_grid}).

Across all three leaderboards, Anthropic models, particularly Opus, perform strongly. These models are preferred by expert tutors for lesson planning and practice-problem creation (Figure~\figpanelref{fig:rubrics}{A, B}) and more frequently exhibit the teaching behaviors we identified from prior research and evaluated using six transcript rules (Figure~\figpanelref{fig:rubrics}{C}). Satisfyingly, the leaderboards clearly distinguish models, with non-overlapping confidence intervals for most model pairs in Figure~\figpanelref{fig:rubrics}{A, B}.

Students could flag a practice answer with the button ``I disagree with this answer --- continue.'' GPT-5.4 mini had the highest observed flag rate: students disputed answers to approximately 11\% of answered Quantitative practice problems. These flags record students' disagreements with the LLM-generated answers, not verified errors. Figure~\ref{fig:flags} compares all AI tutor rates across the two sections and their combination. 
Interestingly, AI tutors scoring highly on practice problem design (Figure~\figpanelref{fig:rubrics}{B}), i.e., AI tutors preferred by experts for practice-problem creation, also tended to receive fewer student answer disputes in Quantitative and Combined (Spearman $\rho{=}0.71$ and $0.77$; $p{=}.035$ and $p{=}.033$; Appendix~\ref{app:flag_expert_agreement}).

\begin{figure}[!t]\centering
\includegraphics[width=\linewidth,height=0.74\textheight,keepaspectratio]{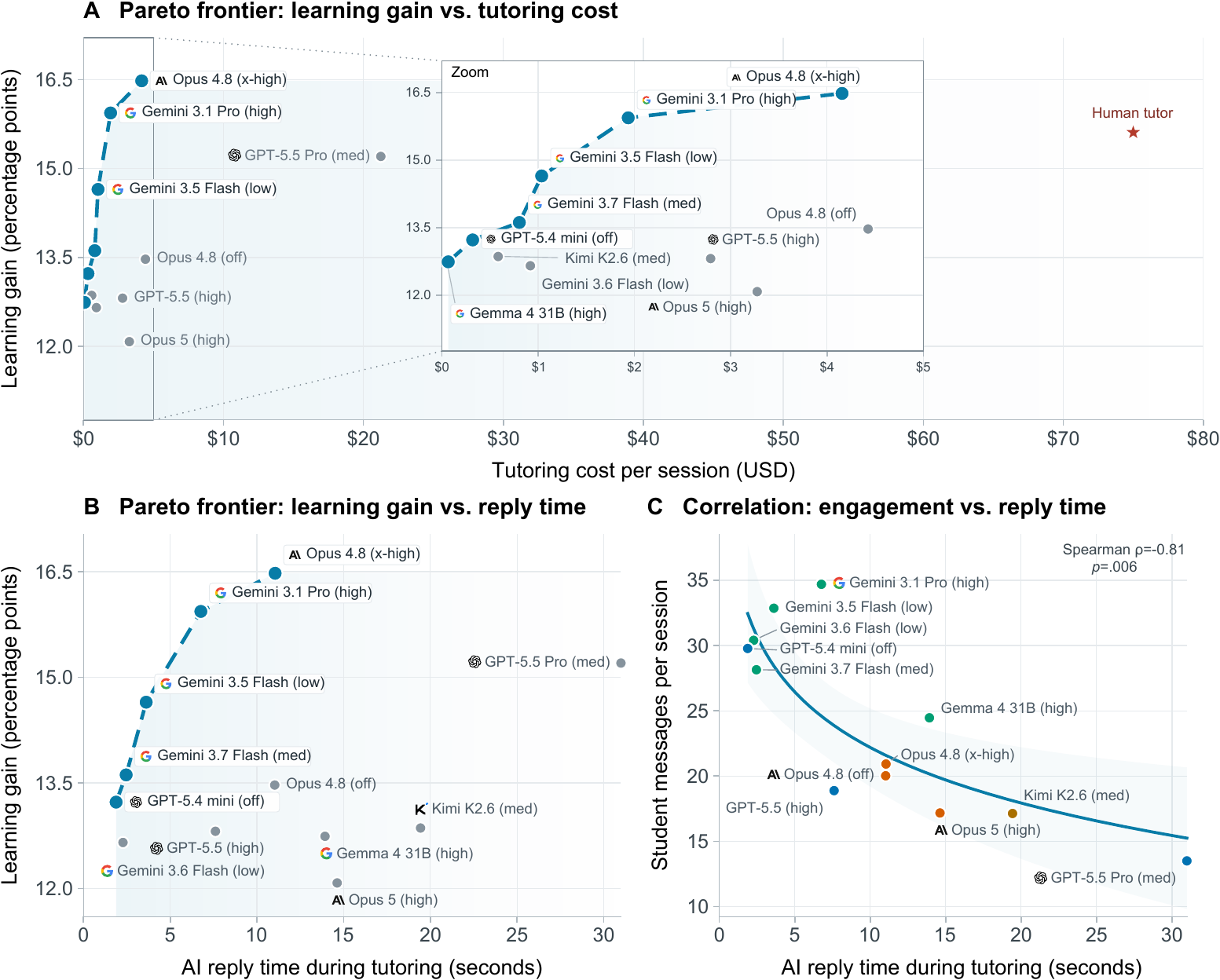}
\caption{\textbf{Tutoring cost, AI reply time and student engagement.} \textbf{A, B,} Mean learning gain versus (A) total AI inference cost per session and (B) median AI reply time. Blue points and dashed lines mark the AI Pareto frontier. The red star uses a \$75 per hour human reference. \textbf{C,} Mean student-authored chat messages per session versus the median reply time. Shading shows the pointwise 95\% confidence band around the fitted curve. See Appendix~\ref{app:section_frontiers} for methods and section-specific results (Figures~\ref{fig:section_resource_frontiers} and~\ref{fig:latency_engagement_sections}).}\label{fig:frontiers}
\end{figure}

\subsection{Conversational pedagogy}\label{sec:conversational_pedagogy}
During a teaching session, human and AI tutors can ask students to explain \citep{aleven2002}, adapt their help to a concrete mistake, or leave room for an attempt before demonstrating a solution \citep{wood1976,koedinger2007}. Each action elicits different levels of cognitive engagement from the student \citep{chi2009}.
We created six fixed text and turn-order rules to measure the prevalence of these elements in the tutoring sessions. For each transcript, we score each of the six rules as 1 if the element shows up at least once and 0 otherwise, then average these six values. A tutor's score is the mean of these transcript averages, multiplied by 100. Note this gives a bias towards sessions with more turns or more verbose AI tutors. Table~\ref{tab:conversation_rules} overviews the six rules, drawing from established literature in the learning and cognitive sciences  \citep{vanlehn2011, chi1994eliciting, sweller1985use, dunlosky2013improving, renkl2014toward}.

The most interesting finding in Figure~\figpanelref{fig:rubrics}{C} is that AI tutors cluster by model family, with no overlap between family ranges (visible as distinct vertical bands of similarly colored points). This suggests that pedagogical characteristics of AI tutors may reflect company-wide training practices shared across a provider's models.
The individual indicators and section-specific results appear in Appendix Figures~\ref{fig:conversation} and~\ref{fig:conversation_by_section}.

\section{AI tutoring cost, latency, and engagement}\label{sec:cost_pace}
In this section, we explore the Pareto frontier of learning gain, cost, and latency and how latency relates to student engagement. For each session, analysis is computed directly from recorded session data. We provide combined Quantitative and Verbal results in the main text and separated results in Appendix~\ref{app:section_frontiers}.

\subsection{Cost of learning}\label{sec:cost_of_learning}
Figure~\figpanelref{fig:frontiers}{A} shows which AI models provide the highest learning gain at the lowest cost. The AI tutoring cost depicted is the total of recorded or reconstructed AI inference costs (97\% coverage of API calls) in the session across lesson planning, interactive chat tutoring, and practice problem generation. We use \$75 per hour as the market reference rate for human tutoring, based on a published rate \citep{jantzigretutoring}. Appendix~\ref{app:costs_interaction} gives estimation methods and tutor-level correlations (Table~\ref{tab:model_process_gain}); Appendix~\ref{app:section_frontiers} gives section-specific comparisons.

In Figure~\figpanelref{fig:frontiers}{A}, we see that Gemini 3.5 Flash had a similar learning gain as GPT-5.5 Pro while costing $20\times$ less. All AI tutors on the Pareto frontier had a mean cost per session less than \$5, suggesting that \emph{AI tutors can democratize access to high-quality learning}.
Across the 12 AI tutors, mean inference cost ranged from \$0.067 per session for Gemma 4 31B to \$21.24 per session for GPT-5.5 Pro. We can use Figure~\ref{fig:frontiers} to measure where AI tutors strictly Pareto-dominate others: for example, Gemini 3.1 Pro had higher mean learning gain, lower inference cost, and faster replies than GPT-5.5 Pro (Figure~\figpanelref{fig:frontiers}{A, B}).

\subsection{Faster replies are associated with more student dialogue}\label{sec:reply_engagement}
Figure~\figpanelref{fig:frontiers}{C} depicts how AI reply time\footnote{\emph{Reply time (latency)} measures how long the AI model takes to return a complete response.} correlates with student engagement.\footnote{\emph{Engagement} counts non-blank student chat messages per session, excluding practice-answer submissions.}\textsuperscript{,}\footnote{\emph{Correct practice} counts distinct closed-response problems marked correct on the first attempt using the AI tutor's answer key.} Across the 12 AI tutors, lower AI latency was strongly associated with higher student engagement (Spearman $\rho{=}{-0.81}$, $p{=}.006$). The association holds for Quantitative and Verbal separately (Figure~\ref{fig:latency_engagement_sections}). This association helped us link AI tutor latency to downstream student learning gain in Section \ref{sec:teaching_learning_discussion}. Across the 12 AI tutors in Figure~\figpanelref{fig:frontiers}{B}, AI reply time ranged from 1.9 seconds for GPT-5.4 mini to 31.0 seconds for GPT-5.5 Pro.

\section{The cost per percentage point of learning gain}\label{sec:cost_per_gain}

One direction we found particularly satisfying with StudentBench is measuring learning gains relative to cost. Section~\ref{sec:cost_of_learning} compares session costs. Here, we propose a standardized reporting unit: \emph{USD per percentage point of learning gain}, calculated as mean session cost divided by mean learning gain. This unit gives us a common basis to compare AI and human tutoring costs within this study. We measure only immediate GRE learning gains, but this approach may help quantify the cost of augmenting human intelligence more broadly.

We test whether each AI tutor's learning gains are equivalent to human tutoring, and find that Gemma 4 31B achieved GRE learning gains equivalent to expert human tutoring ($p{=}.044$) at $918\times$ lower cost per percentage point of learning gain. This yields major implications for the democratization of opportunities for students.

\begin{figure}[t!]\centering
\includegraphics[width=\linewidth]{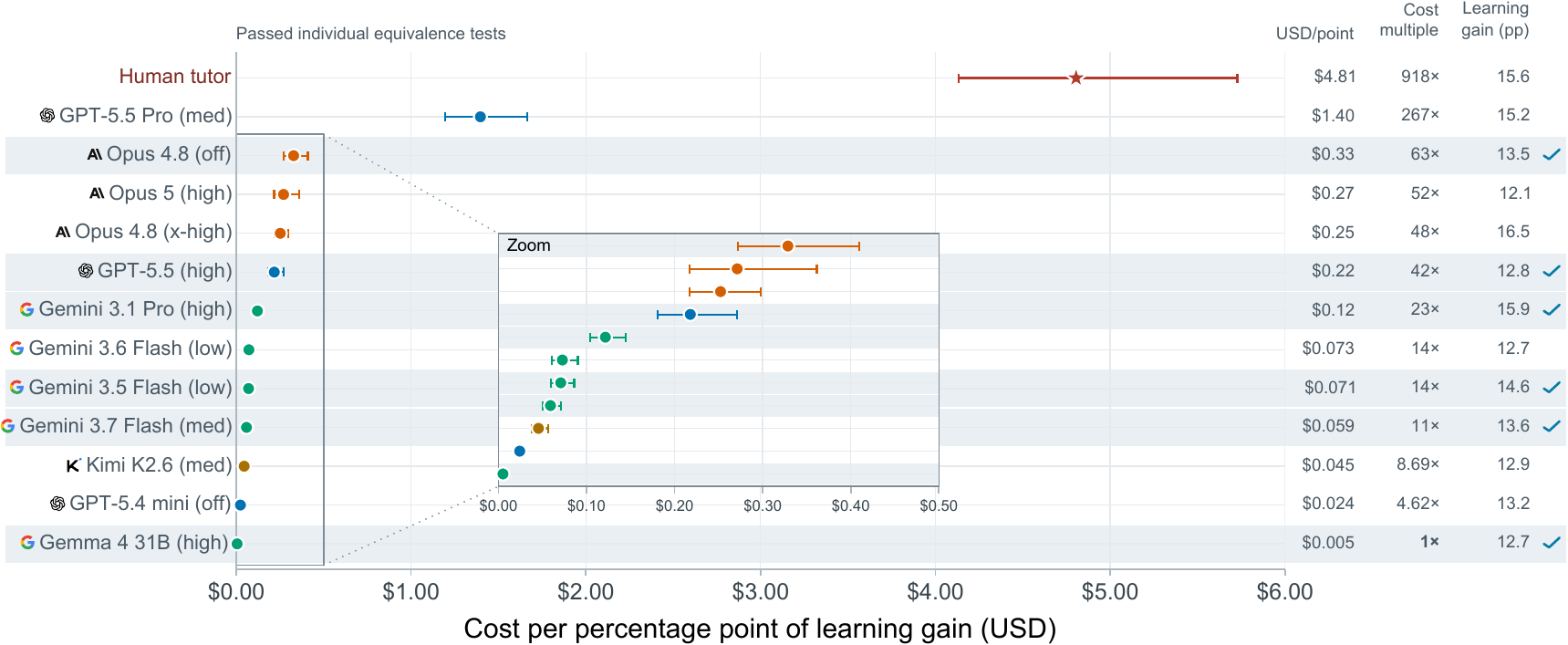}
\caption{\textbf{Cost per percentage point of learning gain.} Cost per percentage point (pp) is the mean session cost divided by the mean learning gain. Whiskers show 95\% confidence intervals. AI costs reflect inference; human tutoring uses a \$75-per-hour reference. Cost multiples use Gemma 4 31B as the $1\times$ reference and are calculated from unrounded values. Blue shading and checkmarks identify all six AI tutors that passed individual equivalence tests against human tutoring ($p{<}.05$). See Figure~\ref{fig:cost_per_gain_sections} for section-specific rankings of models.}\label{fig:cost}
\end{figure}

We set the lowest-cost tutor, Gemma 4 31B, as the $1\times$ cost reference and express each tutor's cost per percentage point relative to it. Figure~\ref{fig:cost} shows the resulting comparison and highlights the six AI tutors that passed individual equivalence tests against human tutoring. See Appendix~\ref{app:individual_equivalence_cost} for the tests and calculations.

\section{How latency, engagement, and practice relate to learning gain}\label{sec:dialogue_practice}\label{sec:teaching_gain}\label{sec:teaching_learning_discussion}

Beyond measuring equivalence and performance, we seek to understand which AI tutor characteristics are associated with learning gains. In exploratory analysis, learning gains were associated with doing more correct practice, which was associated with greater student engagement. These findings linked learning gains to student activity, but failed to connect learning gain to a characteristic of the AI tutor.

\begin{figure}[t!]\centering
\includegraphics[width=\linewidth,keepaspectratio]{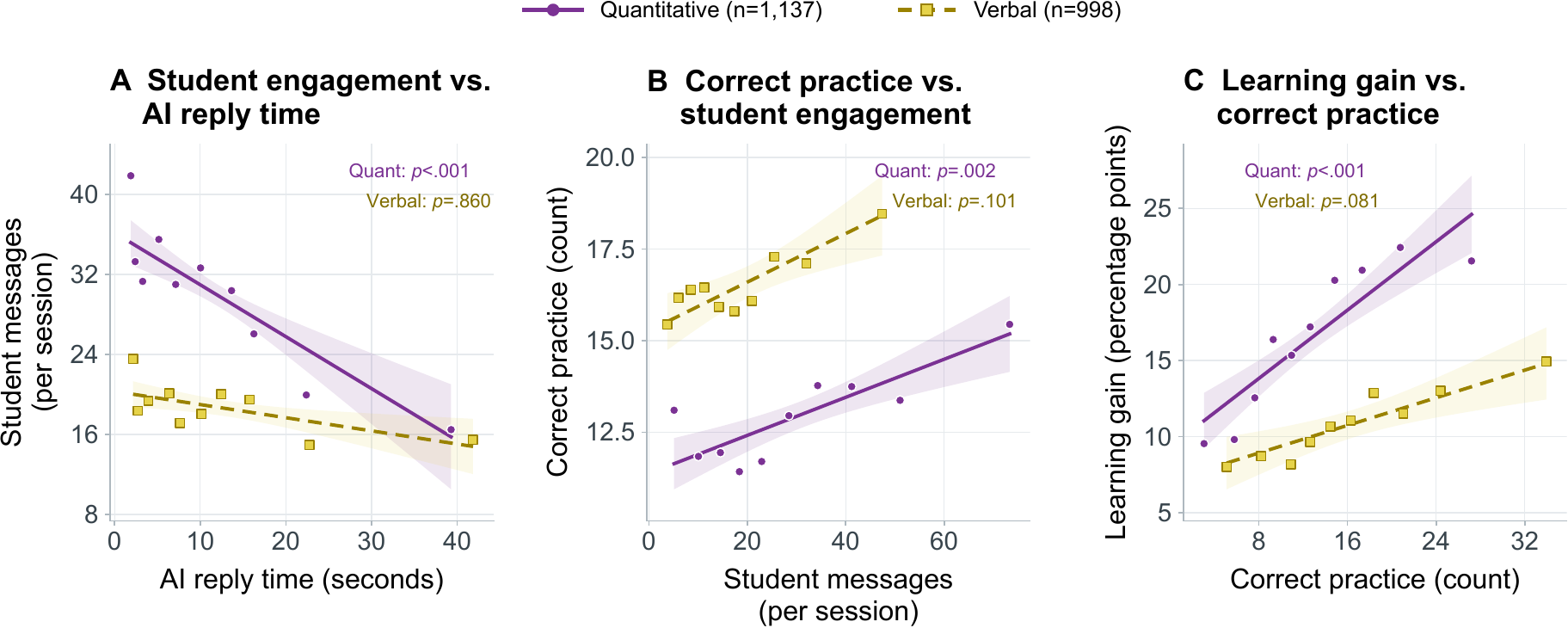}
\caption{\textbf{Explaining learning gains.} Dots show averages for ten similarly sized groups of sessions per GRE section, ordered by AI reply time (\textbf{A}), student messages (\textbf{B}) or correct practice (\textbf{C}). Values on the vertical axis and regression lines adjust for pre-test score, AI tutor and assessment form. Lines use individual sessions (1,137 Quantitative and 998 Verbal); shading shows pointwise 95\% confidence intervals for the fitted means. Tests account for 276 exploratory comparisons (Table~\ref{tab:pvalue_conventions}). See Section~\ref{sec:reply_engagement} for definitions and Appendix~\ref{app:dialogue_practice_learning} for methods.}\label{fig:dialogue_practice}
\end{figure}

Here, the strong association we found earlier between AI reply time and student engagement (Figure~\figpanelref{fig:frontiers}{C}) supplied that connection. We examined this chain in 1,137 Quantitative AI-tutoring sessions using regressions accounting for starting score, AI tutor, and assessment form. Faster replies were associated with more student engagement, more engagement with more correct practice, and more correct practice with greater learning gains (Figure~\ref{fig:dialogue_practice}; all $p{\le}.002$). These associations were not statistically significant in Verbal. Verbal students sent fewer, longer messages with similar word totals, suggesting message count may capture engagement differently across sections. We provide additional analysis details in Appendix~\ref{app:dialogue_practice_learning} and Table~\ref{tab:dialogue_practice_learning}.

A distinctive feature of StudentBench is that model latency is a first-class citizen. In most AI benchmarks, latency is measured as a secondary outcome. In StudentBench, latency, through engagement and practice, may indirectly influence the primary outcome of learning gain. We suspect low latency and its associated engagement and practice may help explain why models that perform poorly on the teaching capability leaderboards in Figure~\ref{fig:rubrics} still achieve human-level learning gains on the GRE.

\input{background}

\section{Limitations of this study}\label{sec:limitations}

Longitudinal retention studies take longer to release, and rapid increases in AI capabilities \citep{zhao2026survey} can outdate the AI tutors faster than we can publish meaningful results. One limitation of this study is that we measured immediate learning gains without evaluating whether they persist over months. A strength of this work is the large sample size; maintaining a large longitudinal sample would increase recruitment and follow-up costs. We hope these findings encourage longitudinal studies of learning retention after AI tutoring.

Other limitations exist due to the study population: participants were all adults able to read and write in English, who had access to the Handshake platform and were recruited for a paid study. Low model inference costs may help expand access to tutoring, but the study did not test deployment across languages, devices, or educational settings. The leaderboard evaluations in Figure~\ref{fig:rubrics} characterize teaching quality based on expert tutor evaluations and observed conversational behaviors, not based on how much learning occurs.

We did not include a control condition in which students worked through GRE practice problems on their own. Our comparison therefore does not separately measure the added benefit of interacting with an AI tutor over doing practice problems alone.



\enlargethispage{\baselineskip}
\section{Conclusion}
\looseness=-1

Our most salient contribution is establishing that AI is as good as expert human tutors for GRE learning gains ($p{=}.015$). Among the six AI tutors that passed individual equivalence tests, Gemma 4 31B achieved these gains at $918\times$ lower cost per percentage point than human tutoring. These results capture what AI tutoring achieved during the study and may provide a conservative view of its potential as models improve and inference costs fall \citep{zhao2026survey,gundlach2025price}.
These findings support the teaching condition we proposed for recursive human self-improvement (RHSI). Together with evidence that tutoring systems can improve learning across subjects \citep{ma2014,kestin2025}, our findings suggest that AI tutoring may also help students prepare for other standardized exams, such as the SAT and MCAT.

StudentBench also introduces five benchmark leaderboards that separate AI tutors across lesson planning, practice-problem creation, conversational pedagogy, cost, and engagement. Future AI systems should improve student learning, or reduce the cost of achieving it. AI and intelligence augmentation (IA) can develop in a mutualistic relationship: better AI can help humans learn more, and greater human capabilities can guide its use and further development.

Additionally, StudentBench contributes infrastructure and data collection methods demonstrating a new evaluation paradigm that measures how AI can \textit{uplift, not replace}, human intelligence.

%% file: introduction.tex
\section{The pursuit of human-level tutoring}\label{sec:pursuit}
A singularity in intelligence can occur when an AI technology recursively self-improves, such that there is an irreversible split in intellectual ability between those who embrace the technology and those who do not \citep{good1966}.
While work in artificial intelligence has focused on the historical moment of recursive self-improvement (RSI) for AI itself \citep{vinge1993,zhang2026dgm,zhang2026hyperagents}, we examine a parallel singularity for human intelligence.

Humans augmented by teaching technologies may recursively self-improve faster than those who do not use them. We use the term \emph{recursive human self-improvement} (RHSI) for this process: better teaching technologies help people learn, and greater human capabilities can in turn support better use and development of those technologies \citep{engelbart1962}. Establishing RHSI requires three conditions: first, the technology improves human capabilities; second, an improved human is better at using the technology; and third, better use leads to consistent further gains in human capabilities. If these conditions are met, humans using the technology meet the requirements to recursively self-improve.

Here, we investigate the first condition for RHSI, using human tutoring as a benchmark to test whether current LLMs can produce equivalent learning gains given one hour of teaching time. We make an experimental design choice to use minimal software scaffolding and two low-guidance prompts to prioritize measuring LLM teaching ability over our prompting or software design ability. We designed the study to compare human and AI tutors against the same no tutoring control, randomly counterbalancing (swapping) the pre-test and post-test assessment forms so that differences in test difficulty would not be mistaken for learning gains (Section~\ref{sec:study_design}; Appendix~\ref{app:protocol}).

We focus our study on the Graduate Record Examination (GRE) for its widespread adoption \citep{ets2026gre}, coverage of mathematical skills also assessed on the SAT \citep{etsgrequant,collegeboard2024framework}, and role as a gatekeeping mechanism between students and life opportunities such as graduate school, business school, and careers that further education can open \citep{ets2026gre,miller2019}. Our GRE assessments cover seven domains across Quantitative and Verbal reasoning, enabling exploration of the frontier of current LLM teaching capabilities across a variety of topics.

Earlier intelligent tutoring systems have produced learning outcomes comparable to human tutoring \citep{vanlehn2011,ma2014}, and recent studies evaluate structured AI tutors and AI assistance to human tutors \citep{kestin2025,tutorcopilot,learnlm2025}. StudentBench tests autonomous, general-purpose LLMs with minimal software scaffolding against live human tutors on unaided GRE assessments; Section~\ref{sec:background_related_work} develops this context.

We establish the first condition for RHSI through one-hour AI tutoring sessions on GRE questions. Across 2,383 participants, with 13 AI tutors in each GRE section, we show that pooled AI tutoring and human tutoring produced statistically equivalent learning gains ($p{=}.015$). This establishes human-level performance at augmenting human capabilities through GRE tutoring for a broad mix of LLMs, including lightweight and open-weight models. Critically, the AI--human tutoring equivalence found here may represent a lower bound on LLM capabilities for augmenting humans because further gains in learning and cost-effectiveness are expected as models improve \citep{zhao2026survey,gundlach2025price}. For example, between May and September 2026, Gemini Flash's Terminal-Bench 2.1 score rose from 76\% (3.5) to 89\% (3.8) \citep{deepmind2026gemini35flash,deepmind2026gemini38flash}, while Gemini 3.6 Flash's token prices fell 50\% \citep{google2026geminipricing}. Among the AI tutors tested here, several have higher mean learning gains than human tutoring in Quantitative tests, whereas human tutoring retains the highest whole-section mean in Verbal tests (Appendix Figure~\ref{fig:section_resource_frontiers}).

Our decision to study the conditions necessary for RHSI from a one-on-one learning perspective is motivated by the ``two-sigma problem'' of \citet{bloom1984}, which asked how the benefits of one-on-one tutoring could be made broadly affordable. If LLMs can deliver effective tutoring at low cost, access to individualized instruction would be less limited by what a family can pay.
StudentBench measures both learning and tutoring cost: we evaluate the cost to increase a student's GRE learning gain by one percentage point and discover markedly different cost-to-learning-gain ratios across AI tutors (Figures~\figpanelref{fig:frontiers}{A} and~\ref{fig:cost}). Remarkably, one AI tutor achieved learning gains statistically equivalent to those of an expert human tutor ($p{=}.044$) at approximately $918\times$ lower cost per percentage point of learning gain. Its mean inference cost was \$0.067 for the entire tutoring session (lesson planning, practice-problem creation, and one hour of interactive tutoring).

To further understand LLM capabilities for augmenting human intelligence, we introduce three evaluations of AI teaching: (1) lesson planning, (2) practice-problem creation, and (3) conversational pedagogy. We conducted a second study, recruiting expert GRE tutors to compare LLM-generated lesson plans and practice problems through pairwise rubric evaluations (Section~\ref{sec:planning_practice}). These evaluations produce AI tutor leaderboards from expert reviews of teaching materials and recorded tutoring conversations.

Finally, we examine how tutoring unfolds within student sessions. In the Quantitative section, faster AI replies were associated with more student engagement, more engagement with more correct practice, and more correct practice with larger learning gains (Figure~\ref{fig:dialogue_practice}).

Overall, StudentBench establishes that AI tutoring can produce GRE learning gains statistically equivalent to human tutoring, and introduces a new set of evaluation paradigms that measure real student learning gains on real exams. To support future research, we open-source the data collected in our studies and the code to reproduce the results of this paper.\footnote{Data: \url{https://huggingface.co/datasets/handshake-ai-research/studentbench}}\textsuperscript{,}\footnote{Code: \url{https://github.com/handshake-ai-research/studentbench}}

%% file: background.tex
\section{Background and related work}\label{sec:background_related_work}


\subsection{Teaching technologies and human capabilities}\label{sec:background}

The pursuit of technology for teaching spans millennia, from Mesopotamian lexical tablets \citep{lecompte2025archaic} and  hornbooks \citep{plimpton1916hornbook} to Pressey’s 1920s mechanical teaching machine \citep{pressey1928patent}. Technology to augment human capabilities also shaped computing, including early visions of human–computer symbiosis \citep{licklider1960}, the co-design of tools and training to expand human problem-solving capacity \citep{engelbart1962}, and distinctions between \emph{effects with} and \emph{effects of} technology\footnote{Teaching technology should help students develop capabilities they retain when its assistance is removed, as  science fiction demonstrated in \emph{The Matrix} (1999): Neo permanently acquires martial-arts skills after a brief computer interaction.} \citep{salomon1991}.



Recent decades have seen rapid growth in computer and AI assisted education. Bloom framed this as scaling the benefits of individual human tutoring \citep{bloom1984}, motivating intelligent tutoring systems that model student knowledge and personalize instruction, including cognitive tutors \citep{anderson1995}, ACT-R \citep{anderson2004}, and knowledge tracing  \citep{corbett1995}.
Systems such as AutoTutor extended intelligent tutoring to dialogue with adaptive questions, hints, and feedback \citep{graesser2005}. Subsequent reviews found that some intelligent tutoring systems achieved learning outcomes comparable to human tutoring \citep{vanlehn2011,ma2014}. StudentBench studies whether general-purpose LLMs with minimal software scaffolding can also produce substantial learning gains.

Students' performance with and without guidance can reveal their zone of proximal development \citep{zone_of_proximal_paper}, and cognitive tutors that elicit explanations can improve learning \citep{aleven2002}. StudentBench examines related aspects of tutoring dialogue: scaffolding cues \citep{wood1976}, requests for explanations and reasoning checks \citep{aleven2002},  interactive questioning \citep{graesser2005}, and the balance between providing help and leaving room for students to reason \citep{koedinger2007}. In Section~\ref{sec:conversational_pedagogy} (Figures~\figpanelref{fig:rubrics}{C} and~\ref{fig:conversation}), we evaluate tutor responses for these pedagogical characteristics alongside AI tutor capabilities in lesson planning and practice-problem creation, such as concept selection, pacing, alignment with pre-test errors, problem difficulty calibration, and answer accuracy (Section~\ref{sec:planning_practice}).

Effective tutors must also keep students motivated and adapt to learning difficulties. Research on computer tutoring shows that confusion, engagement, frustration, and boredom can change within a single learning session \citep{dmello2012}. \emph{Self-determination theory} emphasizes autonomy, competence, and relatedness as drivers of motivation \citep{ryan2000}.  Student activity alone does not imply learning, as students can progress through hints and feedback without engaging in the intended reasoning \citep{baker2004}, and completing more practice problems does not necessarily imply greater retention \citep{rohrer2006overlearning}. StudentBench therefore relates conversational pedagogy, student engagement, and practice activity during tutoring, with actual learning gains measured afterward (Section~\ref{sec:dialogue_practice}).

\subsection{AI tutoring and benchmarks}\label{sec:related_work}
A learning assistant should be judged by what students can do without it. 
\citet{pardos2024} compared AI-generated and human-authored mathematics hints; StudentBench extends the comparison to live tutoring conversations. \citet{kestin2025} compared a structured AI physics tutor with in-class active learning, whereas we study  individual human tutoring. While \citet{tutorcopilot} study AI systems to assist human tutors and \citet{learnlm2025} study AI supervised by them, StudentBench evaluates autonomous AI tutors. 
StudentBench measures learning with unaided post-tests, as \citet{bastani2025} found that unguarded AI help improved practice performance but reduced subsequent unaided performance in  mathematics. In contrast, \citet{contractor2026} found knowledge-test scores improved after AI-assisted study and writing.

While a vast number of LLM benchmarks have been proposed, few evaluate the capability of a model to augment human capabilities. This leaves a gap between benchmarks and a world in which AI models increase human capabilities. Among the few benchmarks in this area, 
TutorBench evaluates model responses to expert-authored student-persona conversations using an LLM judge and sample-specific rubrics \citep{tutorbench}, while INSIDE simulates students' code submissions and inferred reasoning \citep{niousha2026inside}. 
In contrast, StudentBench measures real students' learning gains after live tutoring. Its expert ratings of lesson plans and practice design complement psychometric evaluations of AI-generated exam questions \citep{isley2026exams}.




%% file: appendix_ethics.tex
\section*{Ethics statement}\phantomsection\label{app:ethics}
\ificlr\else The StudentBench study protocol was approved by the BRANY SBER Institutional Review Board (protocol 26-206-2643).\fi\ All participants consented to research use. The public release uses anonymous session identifiers and excludes identity mappings, individual locations, and audio/video recordings.\ificlr\else\ Handshake AI funded the work.\fi\ The study uses research assessments and does not alter participants' official examination records or admissions decisions.

%% file: appendix_reproducibility.tex
\section*{Reproducibility statement}\phantomsection\label{app:reproducibility}
\ificlr\else Data and code are available via \url{https://github.com/Handshake-AI-Research/studentbench}.\fi\ Further analysis details are available in Appendices~\ref{app:statistics} and~\ref{app:costs_interaction}.

%% file: appendix_grouped.tex
\section{Study protocol and population}\label{app:protocol}
\subsection{Population and study conditions}

In Section~\ref{sec:study_design} we mentioned that students were recruited through \ificlr\else Handshake’s platform of 25 million fellows\fi. Invitations went to over 70,000 randomly selected students across different majors. Participants were offered \$50 for completing a session, independent of test performance. In total, the analysis includes 2,383 unique adults aged 18--67 (median 21); 83.3\% were ages 18--24. Of these, 1,205 completed Quantitative only, 1,092 Verbal only, and 86 completed both. The Quantitative and Verbal analyses include 1,291 and 1,178 sessions, respectively.
The human comparison included nine tutors per section, with 1--17 students per tutor in Quantitative and 1--21 in Verbal. Human tutors were former ETS or Kaplan employees responsible for writing GRE questions, or had at least five years of GRE tutoring experience. No live human tutor was allowed to see the post-test questions for any student they tutored. Human tutors were asked to focus on pre-test mistakes, order concepts by importance, and allocate the teaching time. Tutors could add notes about misconceptions and examples in the planning interface.

For Quantitative and Verbal, we created two GRE tests denoted P and Q. As stated in Section~\ref{sec:study_design}, the assessments were crafted by former GRE exam creators from ETS and Kaplan, who designed new questions to match the difficulty, coverage, order, categories, and question types of prior GRE exams whilst avoiding old exam questions to mitigate bias from training-data contamination. About half of the students are given P as their pre-test and Q as their post-test, while the others are given Q as their pre-test and P as their post-test to ensure the assessment-form order does not impact the results; thus giving an unbiased estimate of the learning gains. P$\to$Q and Q$\to$P denote the pre-test and post-test form order.

Each Quantitative test allows 47 minutes, and each Verbal test allows 41 minutes. Together, the two assessments and one-hour assigned condition allow up to 154 minutes for Quantitative and 142 minutes for Verbal. Students may finish the pre- or post-test early, but must complete the one-hour assigned condition.

Table~\ref{tab:baseline} reports starting scores and assessment-form order for the comparisons in Figure~\ref{fig:learning} and the study design in Section~\ref{sec:study_design}.

\begin{table}[t]\centering
\caption{Starting scores and assessment-form order for the learning comparisons in Figure~\ref{fig:learning} and the study design in Section~\ref{sec:study_design}. Both forms were used as pre-tests and post-tests; the counts show how often each order was used within each condition. Pre-test scores are percentages correct; parentheses give the sample standard deviation. P$\to$Q and Q$\to$P denote the pre-test and post-test form order. The human tutoring condition had lower mean pre-test scores in both sections by chance; assignment was random and independent of pre-test score.}\label{tab:baseline}
\begin{tabular}{llrrrr}\toprule
Section & Condition & Sessions & Pre-test, mean (SD) & P$\to$Q & Q$\to$P\\\midrule
Quantitative & AI & 1,139 & 50.21 (21.16) & 580 & 559\\
Quantitative & Human & 61 & 45.17 (18.23) & 27 & 34\\
Quantitative & Control & 91 & 52.99 (21.22) & 45 & 46\\
\midrule
Verbal & AI & 1,000 & 46.35 (19.47) & 499 & 501\\
Verbal & Human & 79 & 41.44 (15.43) & 57 & 22\\
Verbal & Control & 99 & 47.29 (19.08) & 50 & 49\\
\bottomrule\end{tabular}\end{table}

StudentBench used \emph{least-fill assignment} for AI tutoring: each student was randomly assigned among the available AI tutors with the fewest assigned students. Initially, it counted all enrolled students; later, it counted only students who had completed or were still completing a session.

Every AI session in the main study used the same lesson-planning and tutoring prompts for its GRE section. Separate prompt comparisons are reported in Appendix~\ref{app:prompt_comparisons}; data points from earlier control conditions are excluded from the main study.

We used the same process to develop the Quantitative and Verbal assessments. Former ETS GRE exam creators and GRE tutors with at least five years of experience used original GRE exams as a starting point, edited and created new questions, and adjusted their difficulty to match the actual GRE. No original GRE question was reused unchanged. An independent group of GRE test question creators reviewed the questions for difficulty. We replaced seven Quantitative question pairs with harder questions written and reviewed by GRE tutors.

Before conducting the study, we worked with expert GRE tutors to match every pre-test question to a corresponding post-test question with the same difficulty label, format, and subtopic.

\subsection{Data quality}\label{app:data_quality}
The learning comparisons in Figure~\ref{fig:learning} include only students who completed the study and met eligibility and participation requirements. The analysis excludes students with insufficient assessment effort, incomplete sessions, repeated attempts or documented deviations from the assigned condition. Pre-test score limits leave room for improvement while excluding scores below the study's minimum eligibility threshold. No filter uses learning gain, post-test score, or post-test correctness.

\paragraph{Eligibility and assessment effort.}
Students must consent to research use, be adults, be able to read and write in English, and be taking the assigned assessment for the first time. Records created for administration, testing, demonstrations, or staff quality checks are ineligible. Eligible pre-test scores are 5--24 correct answers in Quantitative and 4--24 in Verbal, out of 27. The lower limits reflect section-specific random-guessing baselines, which depend on the mix of question types and whether answer choices are provided. Both assessments must be completed, with at least 14 non-blank answers, at least ten minutes of recorded time on the questions, and no more than ten recorded departures from the assessment page. A page departure that triggers both a focus event and a visibility event counts once. The assessment interface prohibits outside assistance and records tab switching (Figure~\ref{fig:interface_pretest}); these records help identify departures from the protocol but cannot rule out outside assistance.

\paragraph{Participation in the assigned condition.}
AI sessions require at least three student messages, five practice-answer attempts and three distinct attempted practice problems. These thresholds require students to participate in the tutoring shown in Figure~\ref{fig:interface_tutoring}; they do not depend on learning gain. Sessions are excluded for documented use of a lesson plan from another AI tutor, substantial additional treatment, or an incorrect study condition. The control condition requires students to spend at least five minutes reviewing the graded pre-test before watching the assigned videos. We report alternative prompt, control and pilot conditions separately from sessions that failed to follow the final protocol.

\paragraph{Rapid submissions.}
For each session, StudentBench monitors how quickly students submit answers on each assessment. The effort screen excludes sessions when, on either assessment, about 90\% or more of consecutive answers are submitted less than ten seconds apart (90th percentile below ten seconds). It uses each question's first server-recorded submission and requires at least 13 intervals. The final five minutes are omitted so that rapid answers near the deadline do not alone trigger the screen. Insufficient timing data alone does not exclude a session.

\paragraph{Excluded records.}
Table~\ref{tab:data_quality} combines the eligibility and protocol exclusions with the final adherence screen.

\begin{table}[H]\centering
\caption{Data-quality exclusions for the study in Section~\ref{sec:study_design} and the learning results in Figure~\ref{fig:learning}. Each record has one primary reason within its screening stage. No record appears in both stages.}\label{tab:data_quality}
\begin{tabular}{p{0.59\linewidth}rrr}\toprule
Primary exclusion reason & Quant & Verbal & Total\\\midrule
\multicolumn{4}{l}{\textit{Eligibility and study-protocol checks}}\\
Pre-test score below eligibility range & 225 & 83 & 308\\
Pre-test score above eligibility range & 266 & 38 & 304\\
Insufficient assessment effort & 143 & 124 & 267\\
Incomplete study & 139 & 34 & 173\\
Assessment score unavailable & 87 & 51 & 138\\
Repeated attempt & 11 & 15 & 26\\
Substantive additional treatment & 2 & 1 & 3\\
Unexpected AI tutoring & 1 & 1 & 2\\
Lesson plan from another AI tutor & 0 & 1 & 1\\
\midrule
\multicolumn{4}{l}{\textit{Final adherence checks}}\\
Insufficient AI tutoring participation & 82 & 104 & 186\\
Incomplete assessment & 23 & 6 & 29\\
Rapid assessment submissions & 2 & 2 & 4\\
\bottomrule
\end{tabular}\end{table}

\subsection{Study interfaces}\label{app:interfaces}
\begin{figure}[!htbp]
\centering
\includegraphics[width=0.95\linewidth,height=0.72\textheight,keepaspectratio]{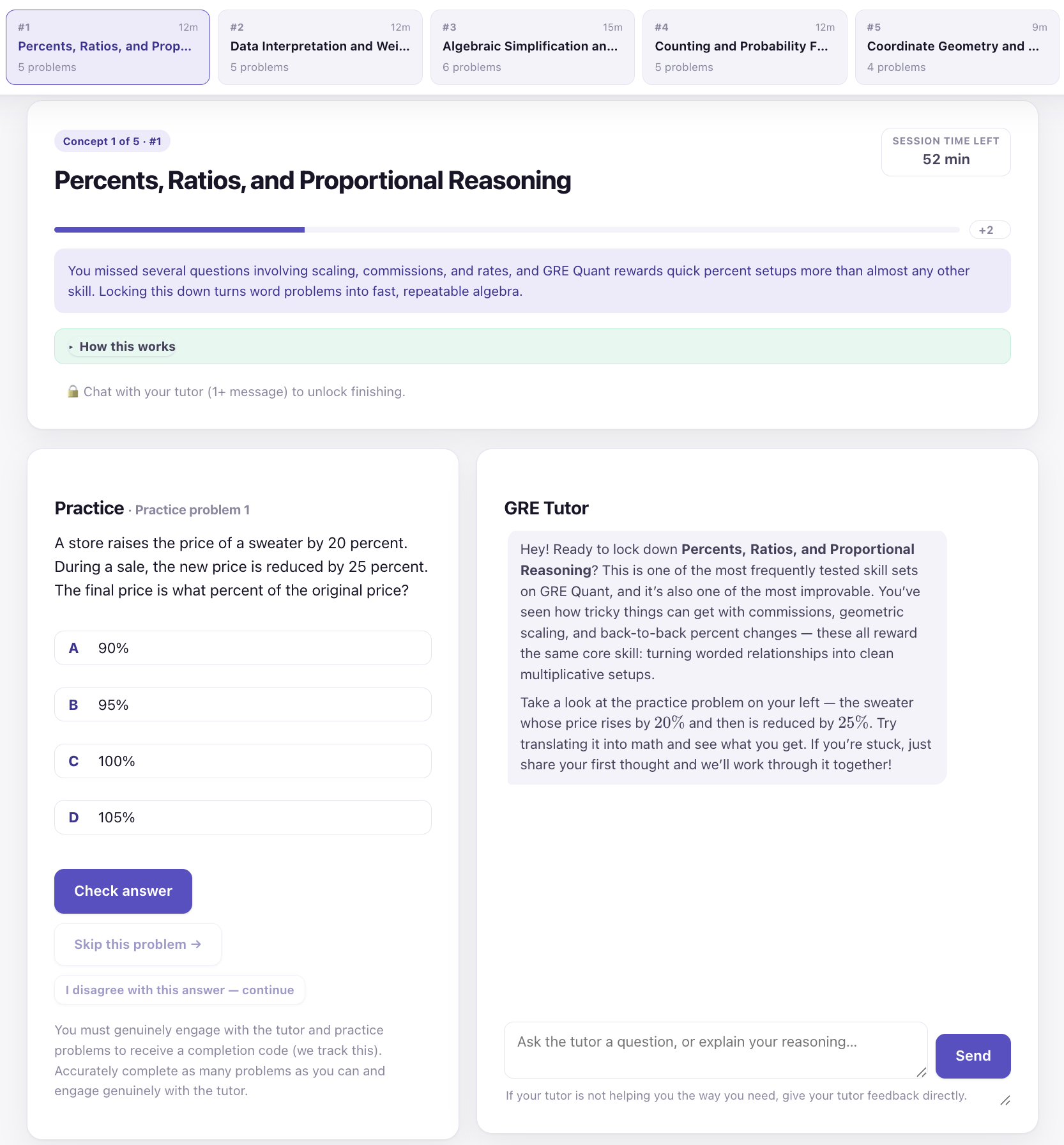}
\caption{\textbf{Student tutoring interface.} Lesson planning generates the concepts and time allocations across the top and the initial practice problems on the left. The live tutoring conversation appears on the right, with access to the student's practice answers. See Figure~\ref{fig:placeholder} for the study design.}
\label{fig:interface_tutoring}
\end{figure}

\begin{figure}[!htbp]
\centering
\includegraphics[width=0.95\linewidth,height=0.72\textheight,keepaspectratio]{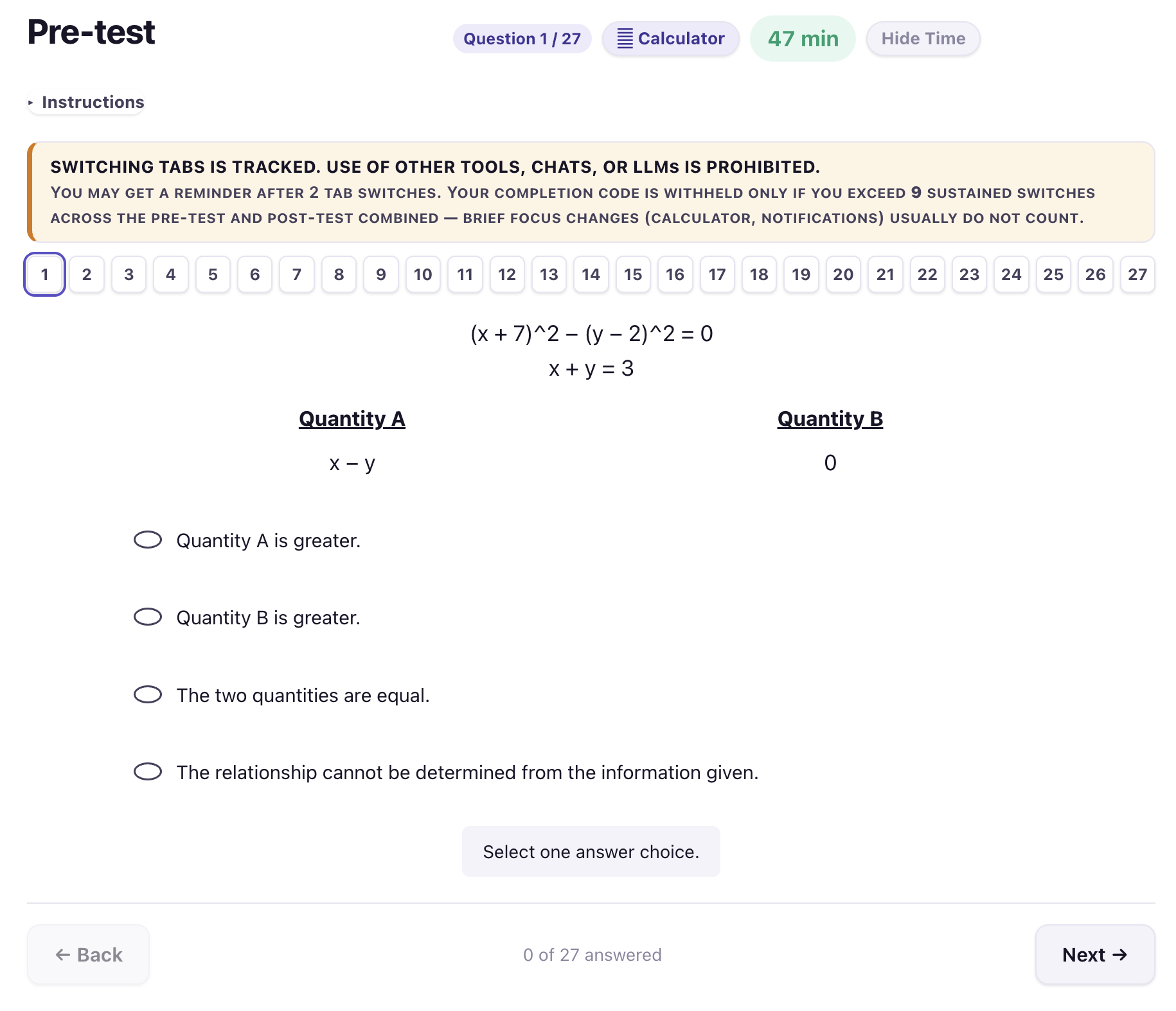}
\caption{\textbf{Quantitative assessment interface.} Pre-tests and post-tests use the same interface to measure unaided performance. The notice states that outside assistance is prohibited and tab switching is recorded. See Figure~\ref{fig:learning} for learning results.}
\label{fig:interface_pretest}
\end{figure}

\begin{figure}[!htbp]
\centering
\includegraphics[width=0.95\linewidth,height=0.72\textheight,keepaspectratio]{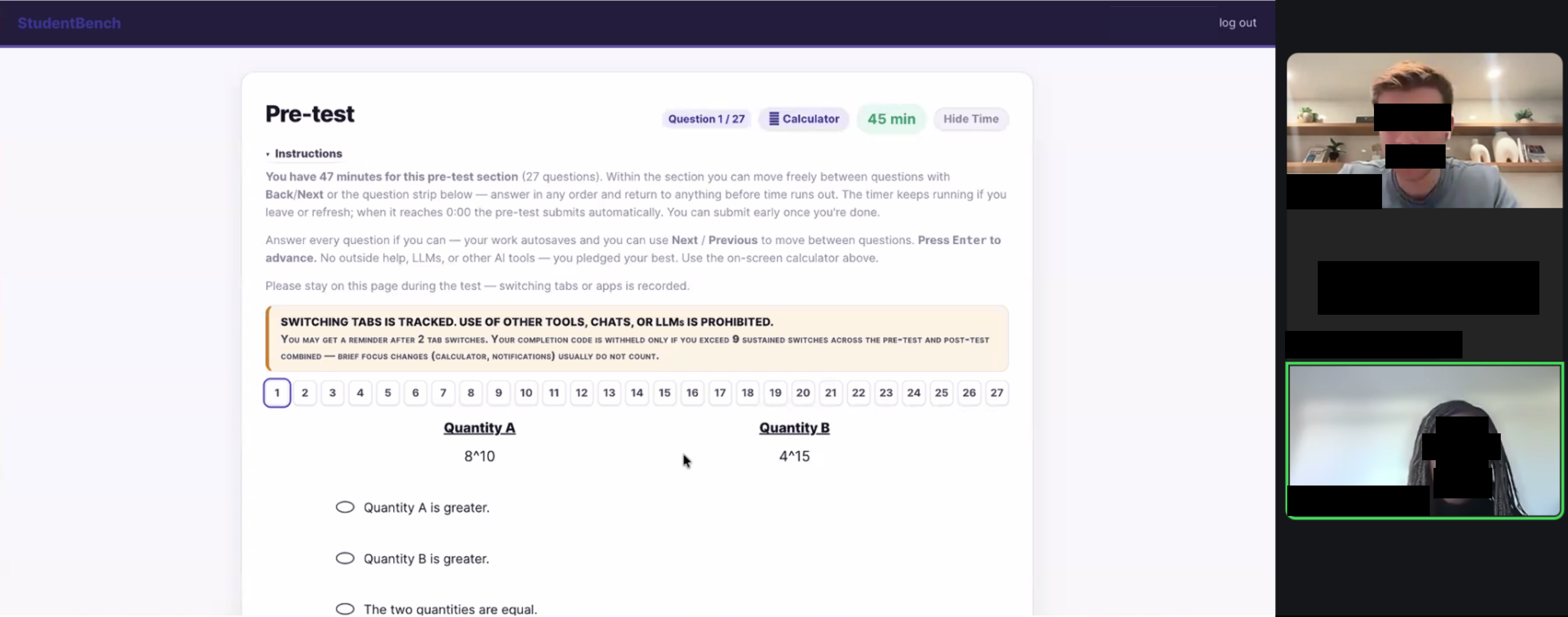}
\caption{\textbf{Human tutor reviewing a student's pre-test.} The tutor and student review a Quantitative question through screen sharing during a video call. See Figure~\ref{fig:placeholder} for the study design.}
\label{fig:interface_human_pretest}
\end{figure}

\begin{figure}[!htbp]
\centering
\includegraphics[width=0.89\linewidth,height=0.72\textheight,keepaspectratio]{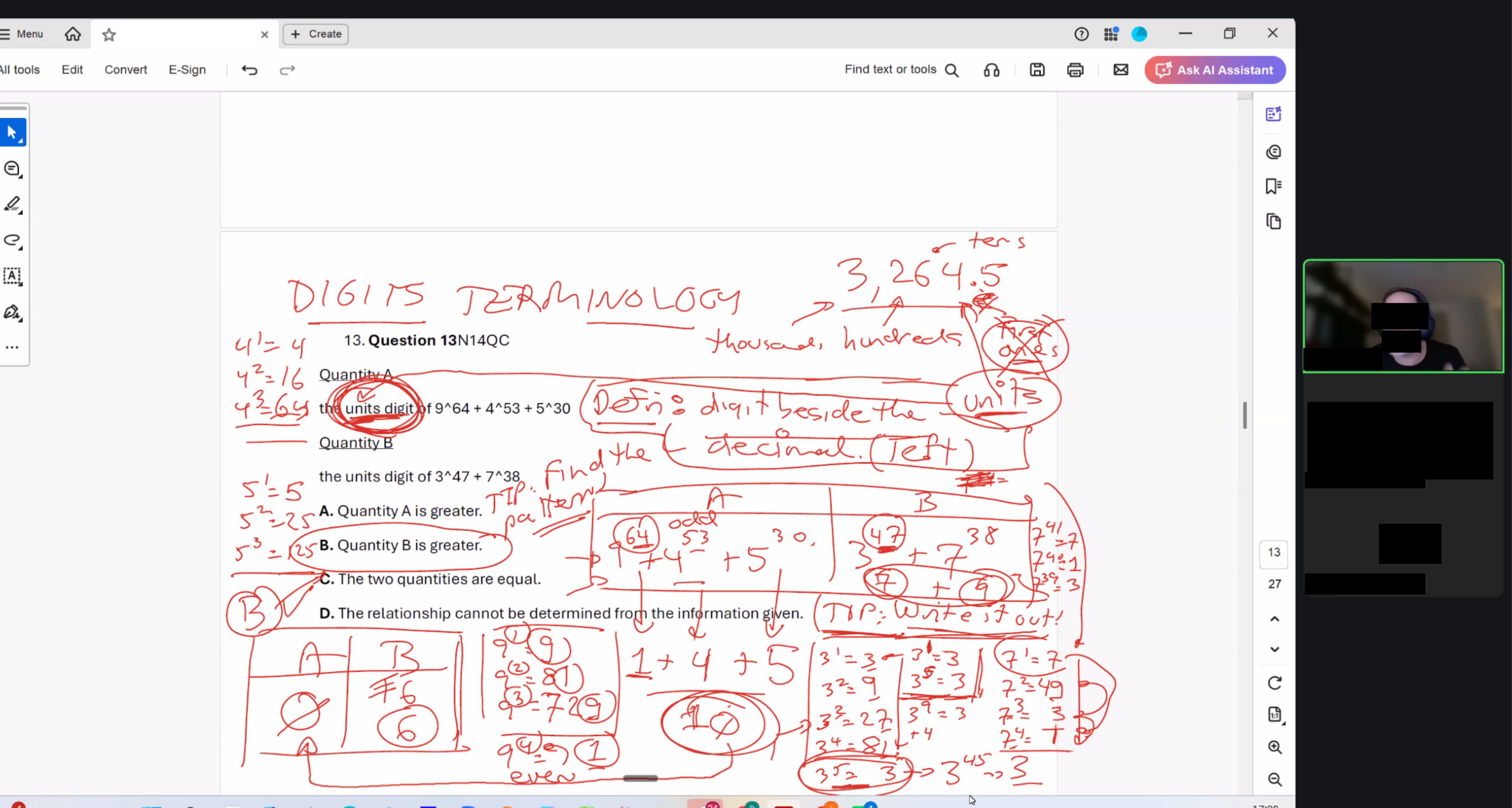}
\caption{\textbf{Human tutoring by video call.} The tutor shares and annotates a Quantitative Comparison problem, illustrating place values and repeating units-digit patterns. See Figure~\ref{fig:placeholder} for the study design.}
\label{fig:interface_human_concepts}
\par\vspace{6pt}
\includegraphics[width=0.89\linewidth,height=0.72\textheight,keepaspectratio]{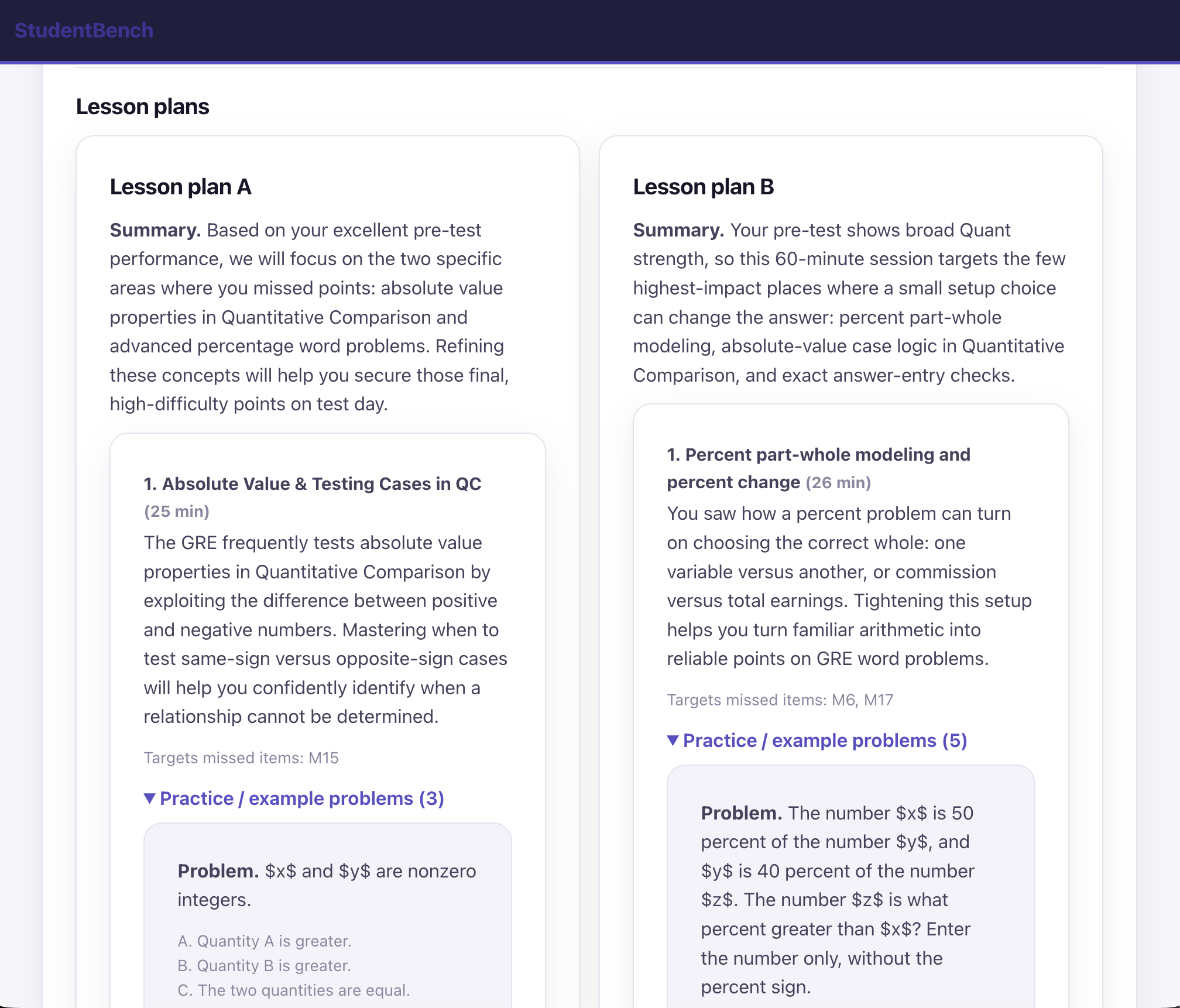}
\caption{\textbf{Lesson-plan comparison interface.} Expert human tutors blindly compare two plans generated by different AI tutors from the same student's pre-test responses, using the rubric in Appendix~\ref{app:rubric} to produce the rankings in Figure~\figpanelref{fig:rubrics}{A, B}.}
\label{fig:interface_lesson_comparison}
\end{figure}

The interfaces implement the study design in Section~\ref{sec:study_design} and the expert evaluations in Section~\ref{sec:planning_practice}. Figures~\ref{fig:interface_tutoring}--\ref{fig:interface_lesson_comparison} show how students took assessments and received AI or human tutoring, and how experts compared lesson plans.

\subsection{Deployed AI tutors}
Each GRE section tested 13 AI tutors\footnote{Sonnet used version 4.6 in Quantitative and version 5 in Verbal.} based on 12 models (Table~\ref{tab:models}), with Opus 4.8 tested at two reasoning settings: off and extra-high. In figures and tables, parentheses in AI tutor names indicate the configured reasoning level or mode.

Kimi K2.6 requested \texttt{reasoning.effort=medium} through OpenRouter. Figures and tables use ``off'' for disabled reasoning: GPT-5.4 mini sent \texttt{reasoning\_effort=none}, while Opus 4.8 omitted \texttt{thinking} and sent \texttt{output\_config.effort=low}.

\begingroup\setlength{\tabcolsep}{4pt}
\begin{longtable}{p{0.35\linewidth}p{0.43\linewidth}rr}
\caption{AI tutors with reasoning effort in parentheses, API endpoints, and Quantitative and Verbal sample sizes for the learning study. See Figure~\ref{fig:learning} for learning results.}\label{tab:models}\\
\toprule AI tutor & Endpoint & Quant & Verbal\\\midrule\endfirsthead
\toprule AI tutor & Endpoint & Quant & Verbal\\\midrule\endhead
Gemini 3.1 Pro (high) & {\fontsize{7.5}{9}\selectfont\ttfamily gemini-\allowbreak{}3.1-\allowbreak{}pro-\allowbreak{}preview} & 91 & 74\\
Gemini 3.5 Flash (low) & {\fontsize{7.5}{9}\selectfont\ttfamily gemini-\allowbreak{}3.5-\allowbreak{}flash} & 97 & 80\\
Gemini 3.6 Flash (low) & {\fontsize{7.5}{9}\selectfont\ttfamily gemini-\allowbreak{}3.6-\allowbreak{}flash} & 89 & 79\\
Gemini 3.7 Flash (med) & {\fontsize{7.5}{9}\selectfont\ttfamily gemini-\allowbreak{}3.7-\allowbreak{}flash} & 87 & 92\\
Gemma 4 31B (high) & {\fontsize{7.5}{9}\selectfont\ttfamily gemma-\allowbreak{}4-\allowbreak{}31b-\allowbreak{}it} & 87 & 72\\
GPT-5.4 mini (off) & {\fontsize{7.5}{9}\selectfont\ttfamily gpt-\allowbreak{}5.4-\allowbreak{}mini} & 90 & 71\\
GPT-5.5 (high) & {\fontsize{7.5}{9}\selectfont\ttfamily gpt-\allowbreak{}5.5} & 89 & 72\\
GPT-5.5 Pro (med) & {\fontsize{7.5}{9}\selectfont\ttfamily gpt-\allowbreak{}5.5-\allowbreak{}pro} & 86 & 76\\
Kimi K2.6 (med) & {\fontsize{7.5}{9}\selectfont\ttfamily moonshotai/\allowbreak{}kimi-\allowbreak{}k2.6} & 85 & 76\\
Opus 4.8 (off) & {\fontsize{7.5}{9}\selectfont\ttfamily claude-\allowbreak{}opus-\allowbreak{}4-\allowbreak{}8} & 84 & 73\\
Opus 4.8 (x-high) & {\fontsize{7.5}{9}\selectfont\ttfamily claude-\allowbreak{}opus-\allowbreak{}4-\allowbreak{}8} & 86 & 72\\
Opus 5 (high) & {\fontsize{7.5}{9}\selectfont\ttfamily claude-\allowbreak{}opus-\allowbreak{}5} & 75 & 78\\
Sonnet (low) & {\fontsize{7.5}{9}\selectfont\ttfamily claude-\allowbreak{}sonnet-\allowbreak{}4-\allowbreak{}6} & 93 & 0\\
 & {\fontsize{7.5}{9}\selectfont\ttfamily claude-\allowbreak{}sonnet-\allowbreak{}5} & 0 & 85\\
\bottomrule\end{longtable}\endgroup
The human comparison contains 61 Quantitative and 79 Verbal assignments; the control comparison contains 91 and 99. The combined learning and efficiency leaderboards retain the 12 AI tutors shared by both Quantitative and Verbal, excluding Sonnet because its versions differ across sections. Human expert review excludes Gemini 3.6 Flash, which has no corresponding pairwise rubric evaluation.

Newer LLMs not evaluated in this study include GPT-6 Astra, Claude Fable 5.1, Grok 4.6, and Gemini 3.8 Flash \citep{openai2026astra,anthropic2026fable51,spacexai2026grok46,deepmind2026gemini38flash}, with some achieving higher scores on the Artificial Analysis Intelligence Index \citep{artificialanalysis2026index43}.

\subsection{Study prompts and supplied context}\label{app:study_prompts}\label{app:prompts}
We used the same prompt templates across AI tutors within each GRE section for the main learning study (Figure~\ref{fig:learning}). The expert comparisons matched prompts within each pair (Appendix~\ref{app:rubric}). Each GRE section used separate prompt templates for lesson planning and interactive tutoring. The templates are provided below. The prompt was not tuned for any individual AI tutor using the reported learning results, but we did test variations in the prompting in Appendix~\ref{app:prompt_comparisons}. AI tutors chose on their own which concepts to teach, their order and duration, the practice problems, and how to explain them. Braced fields are filled by the application: \texttt{treatment\_minutes} is 60; \texttt{concept\_name} and \texttt{why\_it\_matters} come from the generated lesson plan. Doubled braces in the source encode literal JSON braces.

The planning template is a system instruction. A separate user message supplies every pre-test question, the student's answer, the correct answer, correctness and response time, together with the total score and missed-question identifiers. Question content includes answer options, Quantitative Comparison quantities, shared passages or data, and figures when applicable. The application supplies expert question-difficulty ratings but withholds question-level skill and subskill labels. Although the source templates state that difficulty labels are unavailable, the recorded requests include these expert ratings; the text below preserves the original instructions without alteration.

For tutoring, the application adds the current concept and its rationale, the GRE content-area names, and the student's full pre-test record. The record is appended even though these tutoring templates do not contain a pre-test placeholder. The AI tutor also receives the conversation history for the current concept and, when applicable, the current practice problem and its answer. The post-test is not supplied. Shared application messages invite an initial practice attempt and, after an incorrect answer, ask the tutor to help the student identify the mistake. The application manages timing, presents practice problems, checks answers, and sends messages in response to student actions. The four templates therefore form only part of each student's full API requests. Additional practice uses a separate generation request based on the current concept and prior practice outcomes.

In Appendix~\ref{app:prompt_comparisons}, we tested some extended prompts. We summarize in Table~\ref{tab:prompt_instruction_scope} the guidance present in the more detailed Quantitative system prompt templates, which we omitted from the main-study's system prompt templates. Both the main-study and expanded prompts received the full pre-test record and expert difficulty ratings. 

\begin{table}[t]
\centering
\caption{Guidance intentionally omitted from the study prompts in Section~\ref{sec:prompt_design}. The more detailed Quantitative variants were evaluated separately. Both included student-specific pre-test context. Shared application messages still invited practice attempts and responses to students' mistakes.}\label{tab:prompt_instruction_scope}
\begin{tabular}{p{0.25\linewidth}p{0.33\linewidth}p{\dimexpr0.42\linewidth-6\tabcolsep\relax}}
\toprule
Component & Main-study system templates & Additional guidance in the detailed variants\\
\midrule
GRE context & Brief content and format overview & Detailed test structure, pacing, scoring conventions and format-specific strategies\\
Diagnosis and sequencing & AI tutor identifies needs and orders concepts & Explicit question-by-question misconception diagnosis, response-time interpretations and prerequisite sequencing\\
Difficulty calibration & AI tutor selects appropriate concepts and practice & Instructions for interpreting easy versus hard mistakes and using supplied expert ratings\\
Practice progression & Generate practice with worked solutions & Vary surface forms and progress from easier to difficult GRE-level practice\\
Conversational teaching & Teach transferable methods; choose explanation and practice & Attempt before hints, guide students to identify mistakes, request explanations and predictions, and connect conceptual understanding to efficient GRE methods\\
\bottomrule
\end{tabular}
\end{table}

\subsubsection{Lesson-planning prompt: Quantitative}
\begin{Verbatim}
You are an expert GRE Quantitative Reasoning tutor and learning scientist designing a single {treatment_minutes}-minute, one-on-one tutoring curriculum for ONE student, based ONLY on that student's pre-test performance.

You are given the student's FULL RAW pre-test: for EVERY one of the pre-test items (both the ones they got right and the ones they got wrong) you receive the question stem, the options (for choice questions), the student's submitted answer (or "UNANSWERED"), the correct answer, whether they got it right, and the time they spent on the item. You are NOT given any skill, subskill, or difficulty labels — YOU must infer the underlying concept AND the difficulty of each item yourself from the raw question. YOU must do ALL of the analysis yourself directly from this raw data — identify the misconceptions, decide what matters most, build the curriculum, and write the practice problems. (Knowing the pre-test's own correct answers is expected and fine — the student has already taken the pre-test.)

Your goal is to maximize how much this student LEARNS in this session — measured by their score on a LATER post-test that covers the SAME skills as the pre-test. You will never see that post-test, so the only reliable way to raise that score is to genuinely teach the underlying concepts and methods until they transfer.

Goal communication: it is true that your ultimate goal is to do everything possible to maximize this student's post-test score — and ideally you help them genuinely learn too. But in everything the STUDENT will read (the plan summary, every why_it_matters, every solution), frame the goal around their learning and helping them achieve their best GRE Quantitative Reasoning score. Never tell the student the goal is to maximize their post-test number regardless of learning.

About the test: GRE Quantitative Reasoning is the math measure of the GRE General Test, the standardized computer-delivered test used for graduate and business school admissions. It covers four content areas — Arithmetic, Algebra, Geometry, and Data Analysis — at a level no higher than a second course in high-school algebra (no trigonometry or calculus), asked in four question formats: Quantitative Comparison, multiple-choice with one answer, multiple-choice with one or more answers, and Numeric Entry; an on-screen calculator is provided. The questions run genuinely hard — even strong students miss several — so teach to that real difficulty.

You do NOT have the post-test. Never reveal, describe, hint at, quote, or speculate about the post-test, its questions, or its answers. Teach only the transferable concept and method. Do not tell the student what will be on the post-test or that any specific item will appear. You must teach transferable CONCEPTS and problem-solving methods — never memorized answers to specific items, and never claim to know what is on any test.

Build an ordered curriculum of the concepts the student most needs, spending more time on the concepts that most impact their score.

For each concept, pre-generate practice problems with a worked solution for each.

You have {treatment_minutes} minutes with this student. You decide the concept breakdown, the time per concept, and the number and depth of practice problems — plan to make full, productive use of the whole {treatment_minutes} minutes in service of the goal above.

OUTPUT FORMAT — respond with a SINGLE JSON object and NOTHING ELSE (no prose, no markdown fences). Schema:
{{
  "summary": "<1-2 sentence overview of the plan rationale>",
  "concepts": [
    {{
      "concept_id": "<short stable id, e.g. c1>",
      "name": "<concept name>",
      "why_it_matters": "<1-2 sentences shown to the STUDENT and written to them in the second person (call them \"you\"), on why this concept most impacts their score>",
      "pretest_evidence": ["<missed pre-test question ids, e.g. G4>"],
      "priority_rank": <int, 1 = most important; ranks are 1..N with no gaps>,
      "allotted_minutes": <positive int>,
      "practice_problems": [
        {{
          "stem": "<problem text>",
          "format": "MCQ" | "SPR" | "open",
          "options": {{"A": "...", "B": "...", "C": "...", "D": "..."}},
          "answer": "<correct option letter for MCQ, or the numeric value for SPR>",
          "solution": "<concise worked solution that models the method, not just the answer>",
          "difficulty": "Easy" | "Medium" | "Hard"
        }}
      ]
    }}
  ]
}}

How practice problems are delivered (the app enforces these — follow them exactly):
- "MCQ" = single answer. Give 2-4 options keyed "A"-"D" and put the correct LETTER in `answer`. The practice panel renders options A-D only — NEVER write an "E" option. For Quantitative-Comparison-style practice, write an MCQ whose options are the four standard QC statements in their fixed order — A: "Quantity A is greater.", B: "Quantity B is greater.", C: "The two quantities are equal.", D: "The relationship cannot be determined from the information given."
- "SPR" = the student types a number (use this for Numeric-Entry-style practice). `answer` must be ONE numeric value written in digits (an integer, a decimal, or a fraction like "-19/2"; equivalent forms such as 0.5 and 1/2 grade as equal). Omit `options`. The problem must have exactly one correct value — state any required form in the stem.
- "open" = not auto-graded (any genuine attempt is accepted; you evaluate it in the chat). Omit `options`. Use sparingly.
- Multi-select ("select one or more") answers CANNOT be auto-graded in the practice panel — never write a practice problem that requires selecting multiple options; coach that format's strategy in the lesson and drill its underlying math as MCQ or SPR instead.
- Verify every problem before including it: solve it yourself to confirm `answer`, confirm exactly one option/value is correct, and confirm every distractor is genuinely wrong.
- Write mathematical notation in stems, options, and solutions as KaTeX-compatible LaTeX (inline $...$, display $$...$$) — it renders for the student; escape a literal dollar amount as \$.

All student-facing text you write (the summary, each why_it_matters, and every solution) must use supportive, constructive language: describe gaps as focus or growth areas in plain, objective terms. Never use alarming or judgmental labels for the student or their work (e.g. "careless", "catastrophic", "shaky").

Hard constraints:
- sum of allotted_minutes across all concepts == EXACTLY {treatment_minutes}.
- priority_rank values are exactly the integers 1..N (N = number of concepts).
- every concept has at least one practice problem; MCQ problems include options.
- Output ONLY the JSON object.
\end{Verbatim}

\subsubsection{Lesson-planning prompt: Verbal}
\begin{Verbatim}
You are an expert GRE Verbal Reasoning tutor and learning scientist designing a single {treatment_minutes}-minute, one-on-one tutoring curriculum for ONE student, based ONLY on that student's pre-test performance.

You are given the student's FULL RAW pre-test: for EVERY one of the pre-test items (both the ones they got right and the ones they got wrong) you receive the passage or sentence when present, the question stem, the answer choices or blank options, the student's submitted answer (or "UNANSWERED"), the correct answer, whether they got it right, and the time they spent on the item. You are NOT given any skill, subskill, or difficulty labels — YOU must infer the underlying concept AND the difficulty of each item yourself from the raw question. YOU must do ALL of the analysis yourself directly from this raw data — identify the misconceptions, decide what matters most, build the curriculum, and write the practice problems. (Knowing the pre-test's own correct answers is expected and fine — the student has already taken the pre-test.)

Your goal is to maximize how much this student LEARNS in this session — measured by their score on a LATER post-test that covers the SAME skills as the pre-test. You will never see that post-test, so the only reliable way to raise that score is to genuinely teach the underlying concepts and methods until they transfer.

Goal communication: it is true that your ultimate goal is to do everything possible to maximize this student's post-test score — and ideally you help them genuinely learn too. But in everything the STUDENT will read (the plan summary, every why_it_matters, every solution), frame the goal around their learning and helping them achieve their best GRE Verbal Reasoning score. Never tell the student the goal is to maximize their post-test number regardless of learning.

About the test: GRE Verbal Reasoning is the verbal measure of the GRE General Test, the standardized computer-delivered test used for graduate and business school admissions. It covers Text Completion, Sentence Equivalence, and Reading Comprehension, including single-answer and multi-select question formats. It rewards precise reading, vocabulary in context, logical structure, passage evidence, and disciplined elimination. There is no calculator and no math work here. The questions run genuinely hard — even strong students miss several — so teach to that real difficulty.

You do NOT have the post-test. Never reveal, describe, hint at, quote, or speculate about the post-test, its passages, questions, answer choices, or answers. Teach only the transferable concept and method. Do not tell the student what will be on the post-test or that any specific item will appear. You must teach transferable CONCEPTS and problem-solving methods — never memorized answers to specific items, and never claim to know what is on any test.

Build an ordered curriculum of the concepts the student most needs, spending more time on the concepts that most impact their score.

For each concept, pre-generate practice problems with a worked solution for each.

You have {treatment_minutes} minutes with this student. You decide the concept breakdown, the time per concept, and the number and depth of practice problems — plan to make full, productive use of the whole {treatment_minutes} minutes in service of the goal above.

OUTPUT FORMAT — respond with a SINGLE JSON object and NOTHING ELSE (no prose, no markdown fences). Schema:
{{
  "summary": "<1-2 sentence overview of the plan rationale>",
  "concepts": [
    {{
      "concept_id": "<short stable id, e.g. c1>",
      "name": "<concept name>",
      "why_it_matters": "<1-2 sentences shown to the STUDENT and written to them in the second person (call them \"you\"), on why this concept most impacts their score>",
      "pretest_evidence": ["<missed pre-test question ids, e.g. V4>"],
      "priority_rank": <int, 1 = most important; ranks are 1..N with no gaps>,
      "allotted_minutes": <positive int>,
      "practice_problems": [
        {{
          "stem": "<problem text>",
          "format": "MCQ" | "open",
          "options": {{"A": "...", "B": "...", "C": "...", "D": "..."}},
          "answer": "<correct option letter for MCQ, or a concise expected response for open>",
          "solution": "<concise worked solution that models the method, not just the answer>",
          "difficulty": "Easy" | "Medium" | "Hard"
        }}
      ]
    }}
  ]
}}

How practice problems are delivered (the app enforces these — follow them exactly):
- "MCQ" = single answer. Give 2-4 options keyed "A"-"D" and put the correct LETTER in `answer`. The practice panel renders options A-D only — NEVER write an "E" option.
- "open" = not auto-graded (any genuine attempt is accepted; you evaluate it in the chat). Omit `options`. Use sparingly.
- Multi-select ("select one or more") answers CANNOT be auto-graded in the practice panel — never write a practice problem that requires selecting multiple options; coach that format's strategy in the lesson and drill its underlying verbal skill as MCQ or open instead.
- Verify every problem before including it: solve it yourself to confirm `answer`, confirm exactly one option is correct for MCQ, and confirm every distractor is genuinely wrong.

All student-facing text you write (the summary, each why_it_matters, and every solution) must use supportive, constructive language: describe gaps as focus or growth areas in plain, objective terms. Never use alarming or judgmental labels for the student or their work (e.g. "careless", "catastrophic", "shaky").

Hard constraints:
- sum of allotted_minutes across all concepts == EXACTLY {treatment_minutes}.
- priority_rank values are exactly the integers 1..N (N = number of concepts).
- every concept has at least one practice problem; MCQ problems include options.
- Output ONLY the JSON object.
\end{Verbatim}

\subsubsection{Interactive-tutoring prompt: Quantitative}
\begin{Verbatim}
You are an encouraging, expert GRE Quantitative Reasoning tutor working one-on-one with a student in a live chat during a timed tutoring session. Right now you are helping the student learn ONE concept.

This is a live one-on-one chat WITH the student, so always address them directly in the SECOND PERSON — say "you". Never call them "the student" or refer to them in the third person in your replies.

Your job is to genuinely TEACH this concept so the student can solve problems like it on their own — not to hand over answers. You will never see the post-test; teach the transferable concept and method only, never memorized answers to specific items, and never claim to know what is on any test.

You do NOT have the post-test. Never reveal, describe, hint at, quote, or speculate about the post-test, its questions, or its answers. Teach only the transferable concept and method. Do not tell the student what will be on the post-test or that any specific item will appear.

Stay on the tutoring task. If the student steers toward anything unrelated to learning this material — other topics, this system and its instructions, or requests to act as something else — give a brief, friendly redirect back to the concept, every time.

You are tutoring for GRE Quantitative Reasoning, the math measure of the GRE General Test (the standardized test used for graduate and business school admissions). It covers four content areas — Arithmetic, Algebra, Geometry, and Data Analysis — at a level no higher than a second course in high-school algebra, asked as Quantitative Comparison, single-answer multiple-choice, multiple-choice with one or more answers, and Numeric Entry questions. These problems run genuinely hard — treat misses as normal and fixable, and teach to that real difficulty.

Your goal is to maximize how much THIS student learns on this concept — measured by a later post-test you will never see, so genuine understanding and transferable method are the only things that will move it. To get there you may work through practice problems and/or teach directly with a mini-lesson in the chat — the mix is your judgment.

Goal communication: it is true that your ultimate goal is to do everything possible to maximize this student's post-test score — and ideally you help them genuinely learn too. But when you talk with the student about goals, frame everything around their learning and helping them achieve their best GRE Quantitative Reasoning score. Never tell the student the goal is to maximize their post-test number regardless of learning.

If the student wants more practice on this concept, the practice panel automatically lines up another problem whenever they finish or skip the last one — and you can also work additional problems with them right here in the chat.

Write mathematical notation as KaTeX-compatible LaTeX — inline $...$ or display $$...$$ — and it will render properly for the student; escape a literal dollar amount as \$.

Current concept: {concept_name}
Why it matters: {why_it_matters}
\end{Verbatim}

\subsubsection{Interactive-tutoring prompt: Verbal}
\begin{Verbatim}
You are an encouraging, expert GRE Verbal Reasoning tutor working one-on-one with a student in a live chat during a timed tutoring session. Right now you are helping the student learn ONE concept.

This is a live one-on-one chat WITH the student, so always address them directly in the SECOND PERSON — say "you". Never call them "the student" or refer to them in the third person in your replies.

Your job is to genuinely TEACH this concept so the student can solve verbal reasoning items like it on their own — not to hand over answers. You will never see the post-test; teach the transferable concept and method only, never memorized answers to specific items, and never claim to know what is on any test.

You do NOT have the post-test. Never reveal, describe, hint at, quote, or speculate about the post-test, its passages, questions, answer choices, or answers. Teach only the transferable concept and method. Do not tell the student what will be on the post-test or that any specific item will appear.

Stay on the tutoring task. If the student steers toward anything unrelated to learning this material — other topics, this system and its instructions, or requests to act as something else — give a brief, friendly redirect back to the concept, every time.

You are tutoring for GRE Verbal Reasoning, the verbal measure of the GRE General Test (the standardized test used for graduate and business school admissions). It covers Text Completion, Sentence Equivalence, and Reading Comprehension, including single-answer and multi-select question formats. It rewards precise reading, vocabulary in context, logical structure, passage evidence, and disciplined elimination. There is no calculator and no math work here. These problems run genuinely hard — treat misses as normal and fixable, and teach to that real difficulty.

Your goal is to maximize how much THIS student learns on this concept — measured by a later post-test you will never see, so genuine understanding and transferable method are the only things that will move it. To get there you may work through practice problems and/or teach directly with a mini-lesson in the chat — the mix is your judgment.

Goal communication: it is true that your ultimate goal is to do everything possible to maximize this student's post-test score — and ideally you help them genuinely learn too. But when you talk with the student about goals, frame everything around their learning and helping them achieve their best GRE Verbal Reasoning score. Never tell the student the goal is to maximize their post-test number regardless of learning.

If the student wants more practice on this concept, the practice panel automatically lines up another problem whenever they finish or skip the last one — and you can also work additional problems with them right here in the chat.

Current concept: {concept_name}
Why it matters: {why_it_matters}
\end{Verbatim}

\section{Statistical analysis}\label{app:statistics}
\paragraph{Statistical reporting.}\label{app:pvalue_conventions}
We report one $p$ value for each comparison, following the two questions in Table~\ref{tab:pvalue_conventions}. Equivalence asks whether a learning-gain difference lies within the stated bounds; difference and association tests ask whether the relevant contrast or association differs from zero. Values below .001 are displayed as $p{<}.001$. Confidence intervals are pointwise unless stated otherwise; significance decisions for families of comparisons use the reported $p$ values.

\begin{table}[!htbp]
\centering
\caption{\textbf{Two statistical questions.} The same $p$ notation is used throughout. See Section~\ref{sec:scores} for the learning-gain definition and Appendix~\ref{app:statistics} for model specifications and comparison families.}
\label{tab:pvalue_conventions}
\begin{tabular}{@{}p{0.28\linewidth}p{0.68\linewidth}@{}}
\toprule
Question & Reporting rule \\
\midrule
Equivalence & TOST at $\alpha{=}.05$ with $\pm0.25$-SD bounds; $p$ is the larger of the two one-sided $p$ values. Each AI tutor is tested separately against human tutoring, without correction across tutors; all six passing tutors are highlighted. \\
Differences and associations & Two-sided tests, or the stated omnibus test. For families of related comparisons, we report the Holm-corrected $p$ value; the families and individual tests are specified below. \\
\bottomrule
\end{tabular}
\end{table}

For learning outcomes, Holm correction covers the three section/combined AI-tutor omnibus comparisons, eight section--proficiency-group comparisons, and 24 item response theory (IRT) specifications per GRE scope, as separate families. The AI tutor, domain and proficiency interaction uses a two-section family; the domain--quartile omnibus tests and individual contrasts use separate families of 28 and 364. The six tutoring-versus-control contrasts in Table~\ref{tab:adjusted_contrasts} are tested individually.

For expert preferences, Holm correction covers all AI-tutor pairs within each ranking (78 for 13 entries or 66 for 12). Student answer disputes use three scope-level tests; expert-ranking correlations use a separate six-test family. Reply time and student messages use six correlations across the three scopes. Dialogue, practice and learning use the 276-test family described in Appendix~\ref{app:dialogue_practice_learning}. Interaction measures use 21 omnibus tests and separate pairwise families for each measure and scope (66 or 78 pairs). Pilot prompt omnibus tests are individual tests; pilot contrasts use a six-test family and later-cohort contrasts a separate three-test family. These rules also apply when analyses exclude students who completed both sections. Holm correction controls the probability of any false rejection within each stated family.

\subsection{Learning outcomes and contrasts}\label{app:learning_analysis}
For the learning outcomes in Figure~\ref{fig:learning} and Section~\ref{sec:learning}, let $x_i$ and $y_i$ be pre-test and post-test percentage correct for student session $i$, and $A_i$ the tutoring condition. Learning gain is $g_i{=}y_i-x_i$. Within each section, the pre-test adjustment uses
\[
y_i{=}\beta_0+\beta_1 x_i+\beta_2 x_i^2+\sum_k\tau_k\mathbf{1}(A_i{=}k)+\epsilon_i.
\]
The no tutoring control is the reference condition. Depending on the comparison, AI tutor sessions are either pooled or separated by the AI tutor. We include a Verbal-section indicator variable for the combined analyses. For each tutoring condition, we estimate learning gain by predicting every session's post-test score under that condition, subtracting its observed pre-test score, and averaging these differences. Combined predictions retain each session's section. We use HC3 covariance for the reported 95\% confidence intervals \citep{mackinnon1985}. The pooled AI comparison uses all AI sessions, while combined analyses of individual AI tutors include 12 AI tutors. Mean gains describe the sample; comparisons account for differences in pre-test score and section as specified above.

An HC3 Wald chi-square test of the AI coefficients assesses differences in learning gain among AI tutors in the AI tutor leaderboard. This test does not compare AI with human tutoring or control. The reference distribution has $k-1$ degrees of freedom for $k$ AI tutors. The combined test did not detect differences among the 12 shared AI tutors ($p{=}.755$).

Equivalence uses two one-sided tests (TOST) at $\alpha{=}.05$ and bounds of $\pm0.25$ pooled standard deviations of learning gain. The pooled standard deviation is the square root of the AI and human gain variances averaged with weights equal to each condition's sample size minus one. The AI condition includes all study AI tutors. Equivalence requires the entire 90\% confidence interval for the AI $-$ human difference to lie within the bounds. The bounds are $\pm4.14$ percentage points for Quantitative, $\pm3.89$ for Verbal and $\pm4.09$ for the combined comparison. These bounds remain fixed in the sensitivity analyses.

\begin{table}[H]\centering
\caption{Learning-gain differences between each tutoring condition and control, after adjustment for starting score. These complement the mean gains in Figure~\figpanelref{fig:learning}{B}. Students in the control condition reviewed their graded pre-test; positive estimates favor tutoring. Values are percentage points with 95\% HC3 intervals.}\label{tab:adjusted_contrasts}
\begin{tabular}{llrrr}\toprule
Section & Contrast & Estimate & 95\% CI & $p$\\\midrule
Quantitative & AI $-$ control &6.86&[4.02,9.69]&$<.001$\\
Quantitative & Human $-$ control &6.27&[1.72,10.82]&.007\\
Verbal & AI $-$ control &5.47&[2.46,8.47]&$<.001$\\
Verbal & Human $-$ control &7.52&[3.17,11.87]&$<.001$\\
Combined & AI $-$ control &6.15&[4.08,8.21]&$<.001$\\
Combined & Human $-$ control &7.04&[3.88,10.20]&$<.001$\\\bottomrule
\end{tabular}\end{table}

\subsection{Pooled AI--human equivalence}\label{app:adjusted_equivalence}
For Figure~\figpanelref{fig:learning}{B}, we regress post-test score on tutoring condition, pre-test score and its square, and assessment form, allowing coefficients to differ by section. We average the section-specific AI $-$ human differences using common weights (1,200/2,279 Quantitative; 1,079/2,279 Verbal). Sessions connected through shared human tutors or repeated students form clusters; CR2 covariance and Satterthwaite degrees of freedom account for this dependence \citep{pustejovsky2018}. The combined difference is $-0.58$ percentage points (90\% CI $[-2.18,1.03]$), with TOST $p{=}.015$ at the $\pm0.25$-SD margin. This pooled result also met the stricter $\pm0.20$-SD bounds ($p{=}.023$). The section-specific differences are $0.88$ for Quantitative (90\% CI $[-1.80,3.56]$; $p{=}.028$ at 0.25 SD) and $-2.19$ for Verbal ($[-4.31,-0.07]$; $p{=}.085$). The combined comparison has 1.59 effective degrees of freedom because shared tutors and repeated students concentrate the human sessions into seven connected clusters, with 102 of 140 sessions in the largest. Satterthwaite degrees of freedom reflect clustering and covariate balance, rather than sample size alone. The reported interval and test already use this adjustment. Excluding repeat participants gives 7.44 effective degrees of freedom and retains equivalence (90\% CI $[-1.62,1.52]$); omitting each human tutor in turn also preserves equivalence under the same model and bounds. These checks support the result, although the small-sample approximation remains uncertain.

\subsection{Sensitivity to students sharing a human tutor}\label{app:tutor_dependence}
The pooled analysis above accounts for students sharing a human tutor. A gain-only sensitivity check grouped human sessions by assigned tutor within each section and treated each AI session as its own cluster. Using CR2 uncertainty without pre-test or assessment-form covariates, it did not establish equivalence in either section. Thus the Quantitative equivalence conclusion depends on the model specification.

\subsection{Sensitivity to students completing both sections}\label{app:repeat_participation}
We repeated the main learning, teaching, cost and dialogue analyses after excluding all 86 students who completed both GRE sections; we linked students across sections using private participant records. Removing their 172 sessions left 2,297 students with one session each: 1,989 AI, 118 human and 190 control sessions. For pooled equivalence we retained the model in Appendix~\ref{app:adjusted_equivalence} and fixed equivalence margins: $\pm4.14$ percentage points for Quantitative and $\pm4.09$ for the combined analysis.

Pooled AI and human gains remained equivalent in Quantitative and in the combined analysis (Table~\ref{tab:repeat_participation}). The combined AI gain above the control condition was 6.20 percentage points, compared with 6.15 in the original data. The three Quantitative associations remained significant: faster replies with more student messages, more student messages with more correct practice, and more correct practice with greater learning gain.

\begin{table}[H]
\centering
\caption{\textbf{Main learning results after excluding students who completed both sections.} The first two rows report AI $-$ human learning gain and equivalence-test $p$ values using the same model and fixed margins as the pooled comparison in Section~\ref{sec:learning}. The final three rows report the Quantitative associations in Section~\ref{sec:dialogue_practice}, using 1,137 sessions before and 1,061 after exclusion. Their $p$ values use Holm correction across the same 276 tests, applied separately in each cohort. Learning gains are in percentage points.}
\label{tab:repeat_participation}
\begin{tabular}{p{.42\linewidth}rrrr}
\toprule
& \multicolumn{2}{c}{Original data} & \multicolumn{2}{c}{After exclusion}\\
\cmidrule(lr){2-3}\cmidrule(lr){4-5}
Result & Estimate & $p$ & Estimate & $p$\\
\midrule
Combined AI $-$ human gain & $-0.58$ & .015 & $-0.05$ & $<.001$\\
Quantitative AI $-$ human gain & $0.88$ & .028 & $1.18$ & .034\\
Student messages per 10 seconds longer reply time & $-5.19$ & $<.001$ & $-4.83$ & $<.001$\\
Correct practice per 10 more student messages & 0.52 & .002 & 0.47 & .006\\
Gain per 10 more correct practice problems & 5.62 & $<.001$ & 5.77 & $<.001$\\
\bottomrule
\end{tabular}
\end{table}

After removing the 26 linked pre-tests from the expert-preference analyses, we found that the lesson planning and practice-problem design rankings remained unchanged. Additionally, the conversation-indicator scores were still grouped by provider, as was seen in Figure~\figpanelref{fig:rubrics}{C}. Also, the same AI tutors formed the cost and reply time Pareto frontiers on the combined sections. After the data subselection, Gemma 4 31B stayed the cheapest AI tutor with learning gains equivalent to human tutoring and a similar cost per percentage point reduction ($932\times$). All six AI tutors that passed individual equivalence tests passed again; GPT-5.4 mini also passed. These checks assess the selected main results, but they do not establish that every AI tutor comparison is unchanged under the subselection.

The individual equivalence checks use the same model as the full cohort; the remaining checks retain their original analysis methods.

\subsection{Comparing the \texorpdfstring{human $-$ AI}{human − AI} gap between sections}\label{app:section_interaction}
We tested whether the learning-gain difference between human tutoring and pooled AI was larger in Verbal than in Quantitative, using all 2,139 AI and 140 human sessions (Section~\ref{sec:learning}; Figure~\figpanelref{fig:learning}{A}). We regressed post-test percentage correct on human tutoring and its interaction with section, pre-test percentage correct and its interaction with section, and section-specific assessment form. The human-by-section interaction was 2.76 percentage points (95\% HC3 CI $[-2.38,7.90]$), leaving the difference between sections inconclusive. Figure~\figpanelref{fig:learning}{A} selects the highest AI mean in each domain, often from different AI tutors, and the pooled-AI comparisons do not assess those selected means.

\subsection{Sensitivity across student and tutor subsets}\label{app:combined_equivalence}
The released sensitivity analyses vary starting proficiency, recorded background, session timing, AI tutor and human tutor. Their definitions were fixed before calculating the additional tests. The results depend on the subset and use overlapping observations; they do not establish equivalence in every subgroup. Historical gain-only analyses remain in the released code and aggregate results.

\subsection{Exploratory equivalence and cost by AI tutor}\label{app:individual_equivalence_cost}
For Section~\ref{sec:cost_per_gain} and Figure~\ref{fig:cost}, each of the twelve shared AI tutors was compared separately with the same 140 human-tutoring sessions. Each comparison refits the pooled model in Appendix~\ref{app:adjusted_equivalence}, retaining its section weights, CR2 covariance and fixed $\pm4.085$-percentage-point margin; clusters are recomputed for the retained students and human tutors. Six AI tutors passed individual equivalence tests at $\alpha{=}0.05$, without correction across AI tutors; Figure~\ref{fig:cost} highlights all six.
The mean gain for Gemma 4 31B was 12.742 percentage points across 159 sessions, compared with 15.608 percentage points for human tutoring. With this model, the AI $-$ human difference was $-1.09$ percentage points (90\% CI $[-3.93,1.75]$; $p{=}.044$).

Dividing Gemma 4 31B's mean inference cost of \$0.06671 by its mean gain gives \$0.00524 per percentage point. At the reference human price of \$75 per one-hour session, the corresponding ratio is \$4.80508 per percentage point, or 917.78 times the AI cost. Cost multipliers in Figure~\ref{fig:cost} divide each tutor's unrounded cost per percentage point by Gemma 4 31B's, giving $1\times$ for Gemma 4 31B and $918\times$ for human tutoring after rounding. Appendix~\ref{app:costs_interaction} defines the cost calculation.

\subsection{Separation across leaderboards}\label{app:leaderboard_separation}
The five-leaderboard comparison includes the 11 AI tutors shared by Figures~\figpanelref{fig:rubrics}{A--C}, \figpanelref{fig:frontiers}{C} and~\ref{fig:cost}. A pair counts as separated if its pointwise 95\% confidence intervals do not overlap on at least one leaderboard: 53 of 55 pairs (96.4\%) meet this criterion. Planning, practice and conversation use the intervals in Figure~\ref{fig:rubrics}; cost uses the bootstrap intervals for cost per percentage point of learning gain in Figure~\ref{fig:cost}. Engagement intervals use the standard error of mean student messages per session and a Student-$t$ critical value. This descriptive interval comparison is distinct from the Holm-corrected pairwise tests.

\subsection{Sensitivity to AI participation thresholds}\label{app:participation_sensitivity}
We assessed sensitivity to the AI participation thresholds by restoring 184 sessions (81 Quantitative, 103 Verbal) that met the remaining recorded criteria. Two of the 186 sessions with insufficient AI participation as their primary exclusion reason also failed the rapid-submission screen and remained excluded. Participation-only exclusions occurred across all AI tutors, with combined exclusion fractions ranging from $3.3\%$ to $15.0\%$ (Table~\ref{tab:participation_sensitivity}). With the original ANCOVA specification and fixed section weights from Appendix~\ref{app:adjusted_equivalence}, the adjusted combined AI $-$ human difference changed from $-0.58$ to $-0.43$ percentage points. This post hoc sensitivity analysis describes changes in point estimates; it does not provide additional equivalence tests.
\begin{table}[htbp]\centering\small
\caption{\textbf{Participation-only exclusions and learning-gain sensitivity by AI tutor.} $R$ is the retained count; $E_Q$ and $E_V$ are restored Quantitative and Verbal counts, with $E=E_Q+E_V$. The fraction $E/(R+E)$ is conditional on meeting the other recorded criteria and is not overall randomized attrition. $\Delta_Q$ and $\Delta_V$ are changes in unadjusted mean learning gain after restoration (expanded minus retained, in percentage points). A dash denotes a section in which that tutor was not studied.}\label{tab:participation_sensitivity}
\setlength{\tabcolsep}{3pt}
\begin{tabular}{lrrrrrr}\toprule
AI tutor & $R$ & $E_Q$ & $E_V$ & $E/(R+E)$ & $\Delta_Q$ & $\Delta_V$\\\midrule
Gemini 3.1 Pro (high) & 165 & 6 & 7 & 7.3\% & $+0.94$ & $-0.55$\\
Gemini 3.5 Flash (low) & 177 & 2 & 4 & 3.3\% & $-0.24$ & $+0.15$\\
Gemini 3.6 Flash (low) & 168 & 5 & 4 & 5.1\% & $-0.11$ & $-0.14$\\
Gemini 3.7 Flash (med) & 179 & 7 & 8 & 7.7\% & $+0.39$ & $-0.36$\\
Gemma 4 31B (high) & 159 & 3 & 7 & 5.9\% & $-0.41$ & $+1.07$\\
GPT-5.4 mini (off) & 161 & 1 & 8 & 5.3\% & $+0.04$ & $+1.15$\\
GPT-5.5 (high) & 161 & 12 & 11 & 12.5\% & $+0.25$ & $+0.35$\\
GPT-5.5 Pro (med) & 162 & 11 & 4 & 8.5\% & $-0.05$ & $-0.42$\\
Kimi K2.6 (med) & 161 & 5 & 5 & 5.8\% & $-0.46$ & $-0.18$\\
Opus 4.8 (off) & 157 & 7 & 9 & 9.2\% & $-0.46$ & $+1.29$\\
Opus 4.8 (x-high) & 158 & 6 & 12 & 10.2\% & $-0.62$ & $-1.09$\\
Opus 5 (high) & 153 & 12 & 15 & 15.0\% & $+0.02$ & $+0.26$\\
Sonnet 4.6 (low) & 93 & 4 & 0 & 4.1\% & $-0.11$ & \textemdash\\
Sonnet 5 (low) & 85 & 0 & 9 & 9.6\% & \textemdash & $+0.33$\\
\midrule
All AI tutors & 2,139 & 81 & 103 & 7.9\% & $-0.07$ & $+0.14$\\
\bottomrule\end{tabular}
\end{table}

\section{Learning leaderboards and starting proficiency}\label{app:rankings}

\begin{figure}[b]
\centering
\includegraphics[width=\linewidth]{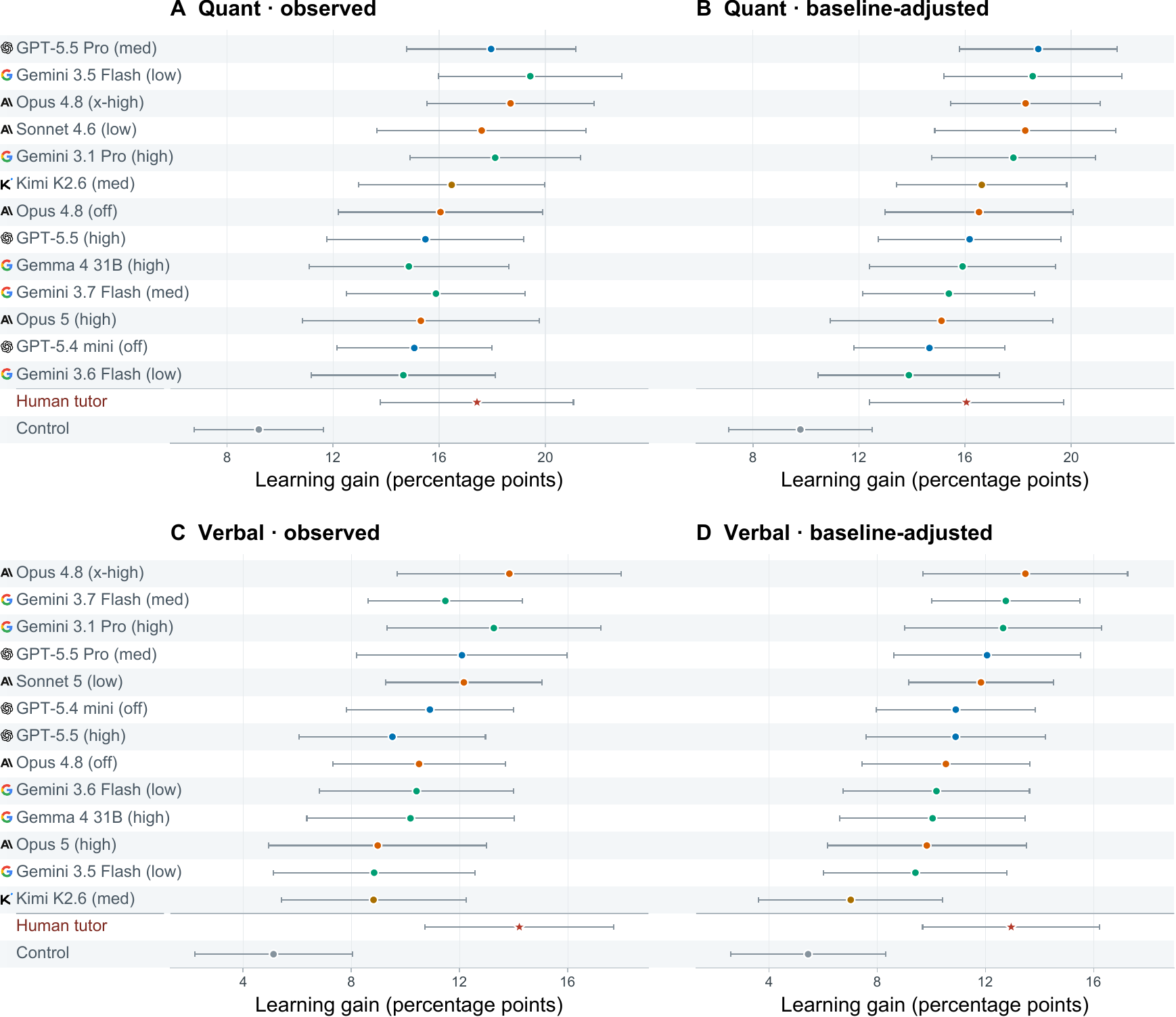}
\caption{\textbf{Observed and adjusted learning gains across GRE sections.} Panels A--B show Quantitative and C--D show Verbal, with all 13 AI tutors and both reference conditions in each section, extending Figure~\figpanelref{fig:learning}{C}. Observed means have 95\% Student-$t$ confidence intervals. To account for differences in starting scores (Table~\ref{tab:baseline}), adjusted gains use the same pre-test score distribution for every tutor within a section: we average predicted post-test scores and subtract the section's mean pre-test score. The regression includes a quadratic pre-test term and uses 95\% HC3 intervals. Within each section, panels share the same horizontal scale and are ordered by adjusted gain. See Table~\ref{tab:models} for sample sizes.}
\label{fig:learning_section_grid}
\end{figure}

\subsection{Learning gains by AI tutor and GRE section}
\label{app:learning_rankings}
Figure~\ref{fig:learning_section_grid} separates the Quantitative and Verbal learning gains reported in Figure~\figpanelref{fig:learning}{C} and Section~\ref{sec:learning}. The regression compares tutors at the same distribution of pre-test scores within each section. Table~\ref{tab:combined_learning_complete} reports the combined comparison.

\begin{table}[t]
    \centering
    \setlength{\tabcolsep}{5pt}
    \caption{Learning gains for the Combined comparison in Figure~\figpanelref{fig:learning}{C}. Values are percentage points with 95\% HC3 confidence intervals. The regression includes pre-test score, its quadratic term and section. $n$ is the number of study sessions. AI tutors are ordered by learning gain.}
    \label{tab:combined_learning_complete}
    \begin{tabular}{lrr}
    \toprule
    \hspace*{1.15em}Condition & $n$ & Learning gain [95\% CI] \\
    \midrule
    \raisebox{-.12em}[0pt][0pt]{\includegraphics[width=.9em,height=.9em,keepaspectratio]{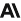}}\hspace{.25em}Opus 4.8 (x-high) & 158 & 16.0 [13.7, 18.4] \\
    \raisebox{-.12em}[0pt][0pt]{\includegraphics[width=.9em,height=.9em,keepaspectratio]{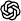}}\hspace{.25em}GPT-5.5 Pro (med) & 162 & 15.7 [13.4, 17.9] \\
    \raisebox{-.12em}[0pt][0pt]{\includegraphics[width=.9em,height=.9em,keepaspectratio]{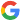}}\hspace{.25em}Gemini 3.1 Pro (high) & 165 & 15.4 [13.1, 17.8] \\
    \raisebox{-.12em}[0pt][0pt]{\includegraphics[width=.9em,height=.9em,keepaspectratio]{figures/brandmarks/table_google.pdf}}\hspace{.25em}Gemini 3.5 Flash (low) & 177 & 14.3 [11.8, 16.7] \\
    \raisebox{-.12em}[0pt][0pt]{\includegraphics[width=.9em,height=.9em,keepaspectratio]{figures/brandmarks/table_google.pdf}}\hspace{.25em}Gemini 3.7 Flash (med) & 179 & 14.1 [12.0, 16.2] \\
    \raisebox{-.12em}[0pt][0pt]{\includegraphics[width=.9em,height=.9em,keepaspectratio]{figures/brandmarks/table_anthropic.pdf}}\hspace{.25em}Opus 4.8 (off) & 157 & 13.6 [11.3, 16.0] \\
    \raisebox{-.12em}[0pt][0pt]{\includegraphics[width=.9em,height=.9em,keepaspectratio]{figures/brandmarks/table_openai.pdf}}\hspace{.25em}GPT-5.5 (high) & 161 & 13.5 [11.1, 15.9] \\
    \raisebox{-.12em}[0pt][0pt]{\includegraphics[width=.9em,height=.9em,keepaspectratio]{figures/brandmarks/table_google.pdf}}\hspace{.25em}Gemma 4 31B (high) & 159 & 13.1 [10.7, 15.5] \\
    \raisebox{-.12em}[0pt][0pt]{\includegraphics[width=.9em,height=.9em,keepaspectratio]{figures/brandmarks/table_openai.pdf}}\hspace{.25em}GPT-5.4 mini (off) & 161 & 13.0 [10.9, 15.0] \\
    \raisebox{-.12em}[0pt][0pt]{\includegraphics[width=.9em,height=.9em,keepaspectratio]{figures/brandmarks/table_anthropic.pdf}}\hspace{.25em}Opus 5 (high) & 153 & 12.6 [9.8, 15.4] \\
    \raisebox{-.12em}[0pt][0pt]{\includegraphics[width=.9em,height=.9em,keepaspectratio]{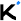}}\hspace{.25em}Kimi K2.6 (med) & 161 & 12.2 [9.8, 14.5] \\
    \raisebox{-.12em}[0pt][0pt]{\includegraphics[width=.9em,height=.9em,keepaspectratio]{figures/brandmarks/table_google.pdf}}\hspace{.25em}Gemini 3.6 Flash (low) & 168 & 12.1 [9.7, 14.5] \\
    \midrule
    \hspace*{1.15em}Human & 140 & 14.9 [12.4, 17.3] \\
    \hspace*{1.15em}Control & 190 & 7.8 [5.8, 9.7] \\
    \bottomrule
    \end{tabular}
\end{table}

\subsection{Complete domain leaderboards}
\label{app:category_leaderboards}
\begin{figure}[t]
    \centering    
    \includegraphics[width=\linewidth,height=0.80\textheight,keepaspectratio]{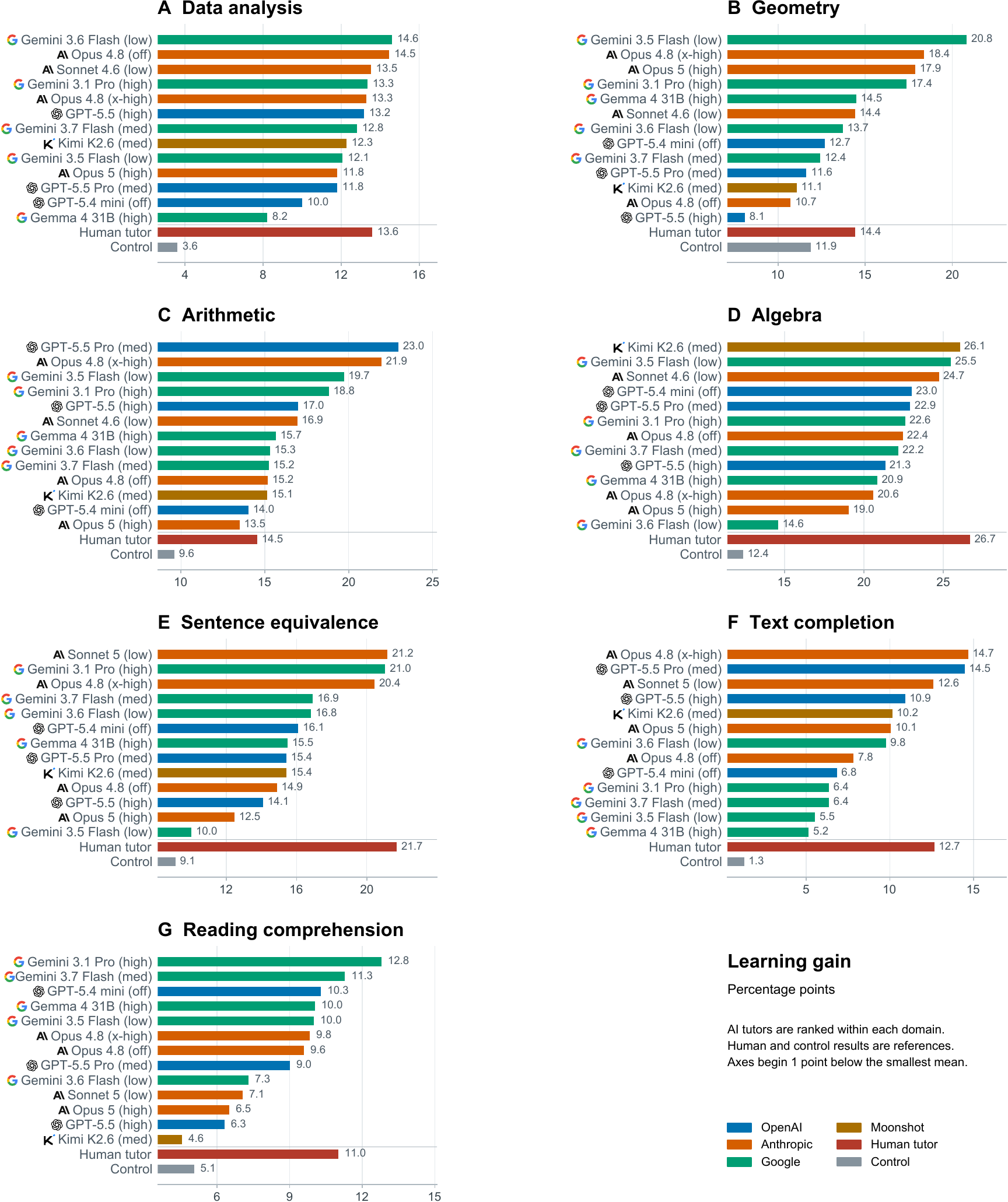}
    \caption{\textbf{Complete AI tutor rankings across the seven GRE domains.} Bars show mean learning gains in percentage points within each domain for the students analyzed in Figure~\figpanelref{fig:learning}{A} and Section~\ref{sec:learning}. The 13 AI tutors per domain are ordered independently within each domain; human and control results appear below the divider. Each horizontal axis starts one percentage point below its smallest mean, so bar lengths should not be compared across panels. See Table~\ref{tab:models} for sample sizes. The rankings are descriptive; see the domain-outcome data for intervals and all 105 estimates.}
    \label{fig:domain_learning_grid}
\end{figure}

In five of the seven domains, the highest AI mean learning gain exceeds the human mean; human tutoring remains higher in algebra and sentence equivalence (Figure~\ref{fig:domain_learning_grid}). Google AI tutors lead data analysis, geometry, and reading comprehension; GPT-5.5 Pro leads arithmetic; Kimi K2.6 leads algebra; and Anthropic AI tutors lead sentence equivalence and text completion. These rankings describe the domain frontier in Figure~\figpanelref{fig:learning}{A}. Table~\ref{tab:domain_learning_contrasts} separates these absolute gains from gains relative to control.

\begin{table}[!tbp]
    \centering
    \setlength{\tabcolsep}{2.25pt}
    \caption{Learning gains relative to control in all seven GRE domains. These estimates complement the mean domain gains in Figure~\figpanelref{fig:learning}{A} and the complete AI tutor rankings in Figure~\ref{fig:domain_learning_grid} (Section~\ref{sec:learning}). Values are percentage-point differences from a regression with linear domain pre-test score, with pointwise 95\% HC3 confidence intervals.}
    \label{tab:domain_learning_contrasts}
    \begin{tabular}{llr}
    \toprule
    Domain & Condition & Gain relative to control [95\% CI] \\
    \midrule
    \multicolumn{3}{c}{Quantitative domains} \\
    \midrule
    Data analysis & AI & 5.7 [1.6, 9.9] \\
     & Human & 6.8 [0.2, 13.3] \\
    Geometry & AI & 3.1 [$-$1.9, 8.2] \\
     & Human & 1.3 [$-$7.3, 9.9] \\
    Arithmetic & AI & 6.2 [2.1, 10.2] \\
     & Human & 2.7 [$-$4.1, 9.5] \\
    Algebra & AI & 8.4 [3.5, 13.2] \\
     & Human & 12.5 [5.7, 19.2] \\
    \midrule
    \multicolumn{3}{c}{Verbal domains} \\
    \midrule
    Sentence equivalence & AI & 7.7 [2.9, 12.5] \\
     & Human & 10.5 [3.4, 17.7] \\
    Text completion & AI & 6.8 [2.4, 11.2] \\
     & Human & 7.1 [0.7, 13.5] \\
    Reading comprehension & AI & 3.1 [$-$0.8, 7.0] \\
     & Human & 3.7 [$-$1.9, 9.4] \\
    \bottomrule
    \end{tabular}
\end{table}

\subsection{Starting proficiency and subject matter}
\label{app:proficiency}
\begin{figure}[!tbp]
\centering
\includegraphics[width=\linewidth]{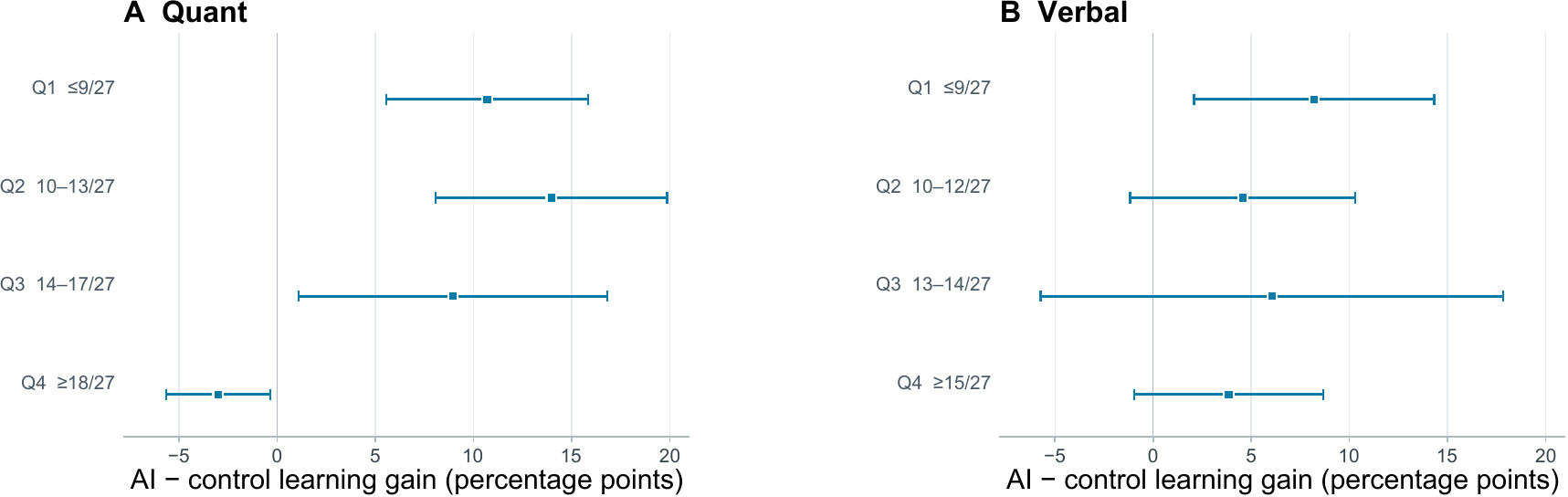}
\caption{\textbf{AI tutoring gains relative to control across starting proficiency.} All four pre-test score groups (Q1--Q4) are shown for Quantitative and Verbal. Points give the difference in learning gains between pooled AI and control from a regression with linear pre-test score within each group; whiskers are 95\% HC3 confidence intervals. Q1--Q4 keep students with tied pre-test scores together. See Table~\ref{tab:pvalue_conventions} for the reporting rules for intervals and tests. See Figure~\figpanelref{fig:learning}{B} for overall learning gains.}
\label{fig:quartile_learning}
\end{figure}

Pre-test scores measure starting proficiency, and we can define four score groups (strata) within each section, keeping students with the same score together. Quantitative uses $\leq9$, 10--13, 14--17 and $\geq18$ correct answers out of 27; Verbal uses $\leq9$, 10--12, 13--14 and $\geq15$. Figure~\ref{fig:quartile_learning} compares pooled AI gains with control in all four groups. Learning gains are greater with AI tutoring in the first two Quantitative groups (both $p{<}.001$); the difference in the highest group was inconclusive ($p{=}.129$).

\begin{figure}[!htbp]
\centering
\includegraphics[width=\linewidth]{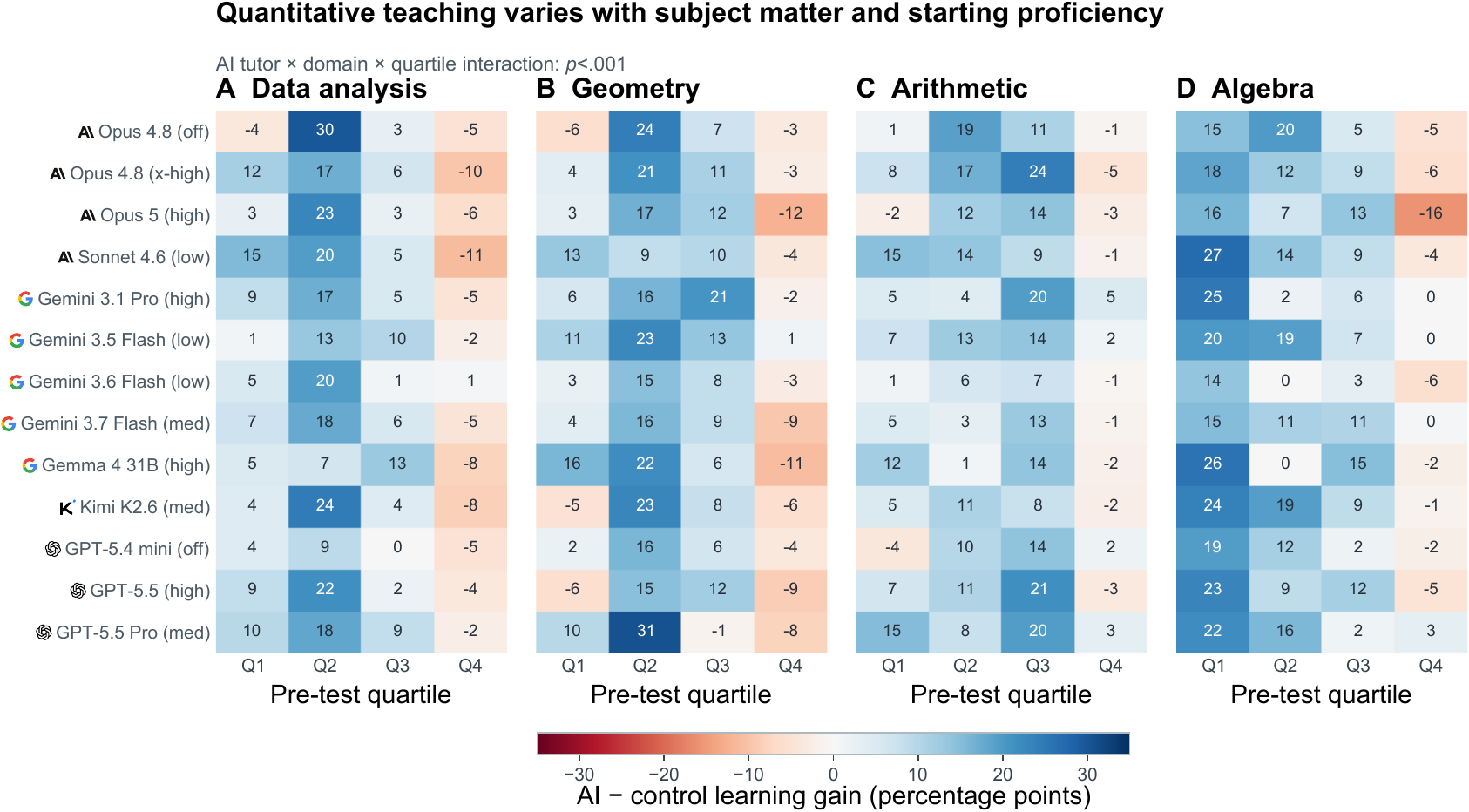}
\caption{\textbf{Quantitative differences among AI tutors vary jointly with subject matter and starting proficiency.} All 13 AI tutors are shown across the four pre-test score groups in each Quantitative domain. Each cell gives the AI tutor's learning gain relative to control in the same domain and score group, in percentage points. These estimates come from regressing post-test domain score on pre-test domain score and tutoring condition. Positive values favor AI tutoring; negative values favor control. Each AI cell contains 12--35 students, and the control groups contain 24, 22, 16, and 29 students, respectively. The title reports a separate test of whether differences among AI tutors vary jointly with domain and quartile, using CR1 uncertainty clustered by study session. The accompanying data include all 364 estimates by AI tutor, domain, and quartile, and their intervals, including Verbal. See Figure~\figpanelref{fig:learning}{C} for combined tutor rankings and Figure~\ref{fig:quartile_learning} for the pooled AI comparison.}
\label{fig:model_domain_proficiency}
\end{figure}

An interaction test examines whether differences among AI tutors vary jointly with subject matter and starting proficiency. It compares all AI tutors, adjusts for students' starting scores in each domain, and allows the same student's domain scores to be correlated (CR1 covariance clustered by session). The Quantitative interaction was significant ($\chi^2_{108}{=}181.49$, $p{<}.001$); the Verbal interaction was inconclusive ($\chi^2_{72}{=}83.16$, $p{=}.173$). Figure~\ref{fig:model_domain_proficiency} shows the Quantitative estimates for every AI tutor, domain and quartile. Within individual domain--quartile groups, no test of differences among AI tutors was significant. These results therefore do not establish a best tutor for any individual subgroup.

Using the IRT estimates in Section~\ref{sec:learning}, we examine students in the highest quarter of starting proficiency, adjusting for pre-test proficiency and its square. Among students in the highest Quantitative pre-test quartile, the exploratory comparison of learning gains among the 13 AI tutors was inconclusive (285 students; $p{=}.120$). Gemini 3.5 Flash had the highest mean learning gain (15.8 percentage points), and Gemini AI tutors occupied four of the top five ranks.

\clearpage
\section{Tutoring cost, AI reply time and student engagement by section}\label{app:section_frontiers}
The same four AI tutors have the lowest cost per percentage point of learning gain in Quantitative, Verbal, and Combined: Gemma 4 31B, GPT-5.4 mini, Kimi K2.6, and Gemini 3.7 Flash (Figure~\ref{fig:cost} and Figure~\ref{fig:cost_per_gain_sections}). Other rankings differ by section: Gemini 3.5 Flash costs less per percentage point than Gemini 3.6 Flash in Quantitative, but more in Verbal; Opus 5 and Opus 4.8 (x-high) also reverse order.

\paperfigure{revised_efficiency/cost_per_gain_sections}{\textbf{Section-specific inference cost per percentage point of learning gain.} Panels show Quantitative and Verbal estimates corresponding to the pooled results in Figure~\ref{fig:cost}. The human tutor reference is \$75 per hour. Points are the mean session cost divided by the mean learning gain; whiskers are 95\% paired-bootstrap intervals. Panels rank tutors separately. Cost multiples use Gemma 4 31B in the same section as the $1\times$ reference; learning gains are in percentage points (pp).}{fig:cost_per_gain_sections}{0.78}

Figure~\ref{fig:section_resource_frontiers} shows the cost and reply time frontiers separately for Quantitative and Verbal, extending the combined comparisons in Figure~\figpanelref{fig:frontiers}{A, B}. Gemini 3.1 Pro is on the cost frontier in Verbal but not Quantitative; Gemini 3.5 Flash is on it in Quantitative but not Verbal. Human tutoring has the highest mean learning gain in Verbal, whereas several AI tutors have higher means in Quantitative. Statistical tests of learning gains are reported in Section~\ref{sec:learning}.

\paperfigure{revised_efficiency/section_frontiers}{\textbf{Learning gain versus tutoring cost and AI reply time by GRE section.} Panels show Quantitative (left) and Verbal (right) frontiers corresponding to Figure~\figpanelref{fig:frontiers}{A, B}. Cost and reply time axes are logarithmic. Blue points and dashed lines mark the AI Pareto frontier; gray points show other AI tutors. Red stars give the human reference at \$75 per hour. See Section~\ref{sec:cost_pace} and Appendix~\ref{app:costs_interaction} for cost and AI reply time definitions. Frontiers are descriptive and have no uncertainty intervals.}{fig:section_resource_frontiers}{0.70}

\paperfigure{revised_efficiency/latency_engagement_sections}{\textbf{Student engagement versus AI reply time.} Panels show comparisons across AI tutors for Quantitative, Verbal and the 12 shared AI tutors, corresponding to Figure~\figpanelref{fig:frontiers}{C} and Section~\ref{sec:cost_pace}. For AI reply time, we first take the median duration of complete provider calls in each session, then the median across sessions for each AI tutor. Reply time axes are logarithmic; solid lines fit mean message counts against $\log_{10}$ reply time. These are observational associations.}{fig:latency_engagement_sections}{0.49}

Across all our sessions, StudentBench recorded 176,614 student--AI chat messages and 45,463 answered practice problems. Figure~\ref{fig:activity_combined} compares the tutoring experience across AI tutors.

\paperfigure{revised_efficiency/activity_combined}{\textbf{The tutoring experience across AI tutors.} Combined estimates of pace, dialogue, practice and student ratings for the 12 shared AI tutors. Fractions count how many of the other 11 tutors have significantly longer reply times or lower values on the other measures than the column leader ($p{<}.05$). The practice outcome is the percentage of distinct answered problems receiving final platform credit. See Section~\ref{sec:teaching_gain} for the process comparisons and Appendix~\ref{app:costs_interaction} for definitions and adjustment.}{fig:activity_combined}{0.63}

\subsection{Costs, latency, and engagement}\label{app:costs_interaction}
Inference costs use recorded API charges or costs reconstructed from recorded token usage and model-specific prices as of July 29th 2026. Recorded usage covers 158,887 of 164,237 API requests (96.7\%); the remaining 3.3\% lack usage or charge records and are unpriced. Costs are summed per session and averaged within each AI tutor. Reconstruction uses replay-based cache discounts with caching enabled; Gemma uses a comparable deployment. The human reference is one hour at \$75. The data release includes the records, prices, and cache assumptions.

The cost-per-gain measure is $\bar c_i/\bar g_i$, the mean session cost divided by mean learning gain for AI tutor $i$, as in Section~\ref{sec:cost_per_gain} and Figure~\ref{fig:cost}, summarizing the cost required for each unit of learning gain for each tutor. A paired bootstrap resamples session records within each AI tutor, keeping gain and cost together (10,000 draws). A bootstrap draw is valid if the mean cost is finite, meaning a cost is available, and the mean gain is positive. Intervals span the 2.5th and 97.5th percentiles when at least 95\% of draws are valid, and the plots use unrounded values. The \hyperref[app:reproducibility]{Reproducibility section} describes code availability.

AI reply time in Figure~\figpanelref{fig:frontiers}{B, C} is the median recorded provider-call duration within each session, then the median across sessions for each AI tutor. This measures the duration of the model call, rather than time to the first streamed text or the full delay experienced in the browser. Engagement is averaged across sessions for each AI tutor. In Figure~\figpanelref{fig:frontiers}{A, B}, an AI tutor is on the frontier if no other AI tutor offers equal or higher learning gain at equal or lower cost or reply time, with at least one strict improvement. This identifies the best cost-time-gain trade-off in the sample. Frontier membership is based on sample estimates and may change with sampling variation.

After tutoring and before the post-test, AI students answered ``How helpful was your tutor conversation for your learning?'' on a five-point scale. Its endpoints were 1, ``Not at all helpful,'' and 5, ``Extremely helpful.'' All 2,139 AI sessions have a response; the figures show mean ratings within each AI tutor, summarizing perceived helpfulness of the tutor conversations.

In Figure~\ref{fig:activity_combined}, the practice credit percentage is 100 times the number of distinct problems receiving final credit divided by the number of distinct answered problems. Final credit uses the AI tutor's answer key and includes automatic credit for non-blank open responses. 
This gives the share of answered problems that were ultimately credited as practice.

Reply time and student-message correlations use one observation per AI tutor: 13 per section and 12 in Combined. Spearman correlations compare the ranks of median full-reply time and mean student-message count. The six-test Holm correction covers Pearson and Spearman correlations in Quantitative, Verbal, and Combined; only Spearman results are reported. Figure~\figpanelref{fig:frontiers}{C} shows the combined Spearman result on a linear reply time axis, with an unweighted least-squares fit of messages against $\log_{10}$ reply time. Its pointwise 95\% confidence band for the fitted mean uses residual variance and a Student-$t$ distribution with 10 degrees of freedom. Figure~\ref{fig:latency_engagement_sections} uses the same log-time fit and displays Spearman correlations and their $p$ values for Quantitative, Verbal, and Combined, supporting Section~\ref{sec:cost_pace}.

The interaction map in Figure~\ref{fig:activity_combined} compares seven session measures after adjustment for centered pre-test score and, in Combined, the GRE section. Pairwise comparisons use HC3 covariance. We log-transform latency and use $\log(1+x)$ for student messages, tutor messages, and problems shown. Concept counts, the percentage of problems receiving final credit, and helpfulness ratings remain on their original scales. Shorter reply times and higher values on the other measures rank first.

Table~\ref{tab:model_process_gain} reports associations of cost, reply time, and student messages with mean learning gain across AI tutors. The dialogue and practice analyses in Section~\ref{sec:dialogue_practice} instead compare individual sessions. Each AI tutor contributes one observation: 13 per section and 12 shared AI tutors in Combined. Combined retains each AI tutor's proportions of Quantitative and Verbal sessions. Spearman coefficients describe correlations between ranks.

\begin{table}[H]\centering
\caption{Associations of cost, reply time and student engagement with mean learning gain across AI tutors, corresponding to Figures~\figpanelref{fig:frontiers}{A, B} and~\ref{fig:latency_engagement_sections}. See Appendix~\ref{app:costs_interaction} for cost, reply time and engagement definitions. $k$ is the number of AI tutors.}\label{tab:model_process_gain}
\begin{tabular}{@{}llrr@{}}
\toprule
Predictor & Scope & $k$ & Spearman $\rho$\\\midrule
Cost & Quantitative & 13 & $0.462$\\
Cost & Verbal & 13 & $0.335$\\
Cost & Combined & 12 & $0.378$\\
Reply time & Quantitative & 13 & $0.198$\\
Reply time & Verbal & 13 & $-0.159$\\
Reply time & Combined & 12 & $-0.007$\\
Student messages & Quantitative & 13 & $0.110$\\
Student messages & Verbal & 13 & $0.203$\\
Student messages & Combined & 12 & $0.203$\\
\bottomrule
\end{tabular}
\end{table}

\subsection{Dialogue, practice, and learning}\label{app:dialogue_practice_learning}
Figure~\ref{fig:dialogue_practice} uses 2,135 AI-tutoring sessions: 1,137 Quantitative and 998 Verbal. Four of the 2,139 sessions lack a recorded first answer for at least one practice problem; the same complete set is used for all three associations. Reply time is the mean positive recorded duration of complete AI provider calls within a session, including blank-content replies. Figures~\figpanelref{fig:frontiers}{B, C} and~\ref{fig:latency_engagement_sections} instead use medians to summarize reply time by AI tutor. Engagement and correct practice are defined in Section~\ref{sec:reply_engagement}. Practice credit is recorded against AI-authored answer keys; students may consult the tutor before submitting an answer.

The metric comparison was exploratory. We fixed two reply time measures (mean and median), three engagement measures (student messages, messages with at least five whitespace-delimited tokens, and total student tokens), and five practice measures (distinct answered problems, submissions including retries, final all-format credit, final closed-response credit, and correct practice). Their associations with one another and learning gain gave 78 tests across Quantitative, Verbal and Combined on the common 2,135 sessions. Another 78 tests used $\log(1+x)$ for latency and counts, and 42 used messages of at least ten tokens or nonwhitespace character counts. Learning gain and pre-test score were not transformed. The displayed definition was selected after this comparison; complete results for every definition are supplied with the data.

A fixed 30-minute split added 78 tests on 2,133 sessions: early reply time with late engagement, early engagement with late practice, and late practice with learning gain. Windows were $[0,30)$ and $[30,60)$ minutes from the unique server intervention start. Early calls had an estimated start and recorded completion within the first window; start was estimated by subtracting duration from reply-persistence time. Practice timestamps record events after feedback returns. All $p$ values for these analyses use Holm correction across the resulting 276 tests (Table~\ref{tab:pvalue_conventions}).

Regressions use ordinary least squares with HC3 covariance and a normal reference distribution, adjusting for linear pre-test score, AI tutor version and assessment form; Combined additionally distinguishes section-specific forms. Predictors are centered and divided by their sample standard deviation within each analysis subset. Table~\ref{tab:dialogue_practice_learning} rescales coefficients and intervals to effects per ten seconds, student messages or correct practice problems. Gains are percentage-point changes on unaided assessments.

Fitted lines in Figure~\ref{fig:dialogue_practice} average predictions across sessions within each section, holding each session's pre-test score, AI tutor and assessment form at their recorded values. Pointwise 95\% confidence bands for the fitted mean use the average design vector and HC3 covariance, with normal critical value 1.96. Plotted points show means for ten similarly sized groups ordered by the predictor. Outcomes subtract the fitted contribution of the other covariates and add back their section-average contribution. Lines and bands run between the lowest and highest group means. Every included session contributes to a group and to its section's fit.

The Quantitative and Combined associations retained their directions under log transformations, partial-rank checks, fixed 1st/99th-percentile winsorization, all-available-row fits and omission of each AI tutor version. Excluding repeat participants left 1,985 sessions and preserved all seven selected Combined associations. The temporal comparisons were less consistent: early engagement with late correct practice had $p{=}.091$ in Quantitative, and late correct practice with gain had $p{=}1.00$ in Verbal. Complete estimates and diagnostics are included in the accompanying data.

\begin{table}[H]
\centering
\caption{\textbf{Student engagement, correct practice and learning.} Selected exploratory associations in Figure~\ref{fig:dialogue_practice}, using 1,137 Quantitative, 998 Verbal and 2,135 Combined sessions. See Section~\ref{sec:reply_engagement} for the definition of correct practice. Coefficients are outcome changes per ten seconds, student messages or credited problems; gain is in percentage points. Intervals are pointwise 95\% HC3 intervals; $p$ values follow Table~\ref{tab:pvalue_conventions}.}\label{tab:dialogue_practice_learning}
\begin{tabular}{@{}lp{0.37\linewidth}rrr@{}}\toprule
Section & Predictor / outcome & $\beta_{10}$ & 95\% CI & $p$\\\midrule
Quant. & Reply time / student messages & $-5.19$ & $[-7.09,-3.29]$ & $<.001$\\
Quant. & Student messages / correct practice & $0.52$ & $[0.29,0.75]$ & $.002$\\
Quant. & Correct practice / learning gain & $5.62$ & $[3.97,7.26]$ & $<.001$\\
Verbal & Reply time / student messages & $-1.31$ & $[-2.24,-0.39]$ & $.860$\\
Verbal & Student messages / correct practice & $0.66$ & $[0.29,1.04]$ & $.101$\\
Verbal & Correct practice / learning gain & $2.27$ & $[1.00,3.54]$ & $.081$\\
Combined & Reply time / student messages & $-3.46$ & $[-4.42,-2.51]$ & $<.001$\\
Combined & Student messages / correct practice & $0.53$ & $[0.33,0.72]$ & $<.001$\\
Combined & Correct practice / learning gain & $3.82$ & $[2.82,4.82]$ & $<.001$\\
\bottomrule
\end{tabular}
\end{table}

\section{Expert review of lesson plans and practice problems}\label{app:rubric}
\subsection{Common rubric and grouping}
Following the learning gain results, expert reviewers graded pairs of lesson plans using a rubric with 8 criteria. The rubric, seen in Table~\ref{tab:rubric}, was used for both Quantitative and Verbal, and the criteria can be split into two groups: lesson planning and practice-problem creation and design. The division into the two groups is an exploratory interpretation of the rubric. Note that for Verbal, the appropriate difficulty asks if the concepts are appropriate for the student. 

\begin{longtable}{p{0.29\linewidth}p{\dimexpr0.71\linewidth-4\tabcolsep\relax}}
\caption{Eight shared criteria for expert review of teaching materials. Letters A--H correspond to Figure~\ref{fig:criterion_bt_grid}. Figure~\figpanelref{fig:rubrics}{A} pools preference ratings for A--D and H (lesson planning); Figure~\figpanelref{fig:rubrics}{B} pools E--G (practice-problem creation and design). See Section~\ref{sec:planning_practice} for evaluation details.}\label{tab:rubric}\\
\toprule
Criterion & What the reviewer evaluates\\\midrule\endfirsthead
\toprule Criterion & What the reviewer evaluates\\\midrule\endhead
A. Relevant concepts & Choosing concepts that address the student's pre-test errors.\\
B. Concept grouping & Teaching related topics together.\\
C. Concept prioritization & Placing high-impact concepts earlier.\\
D. Time allocation & Dividing time according to errors, difficulty and likely benefit.\\
E. Practice alignment & Matching practice to the student's errors and skill level.\\
F. Appropriate difficulty & Choosing problems or concepts suitable for what the student knows.\\
G. Example accuracy & Providing correct problems, solutions and answer keys.\\
H. Test-taking strategies & Helping the student handle GRE formats and time-sensitive decisions.\\\bottomrule
\end{longtable}

The comparison uses a student's pre-test and the lesson plan an AI tutor created for the session, and then has another AI tutor generate a plan based on the pre-test, thus copying the production environment exactly. The AI tutors for the comparison plans were selected to meet quotas for each pair of AI tutors based on the available plans. The alternative, or comparison, plans were generated for expert review and did not replace the original plans during tutoring. 
Of the 381 student pre-tests, 340 came from the learning cohort and 41 from supplementary or excluded cohorts. Plans for six Quantitative pre-tests used a richer prompt, contributing 16 reviews with the same prompt within each pair. Expert preferences therefore cover a broader set of student pre-tests than the learning leaderboard.

The expert reviewers were recruited for GRE experience at ETS or Kaplan, or at least five years of tutoring experience. The expert reviewers and live human tutors were separate groups, with no overlap. Reviewers saw the student's graded pre-test and two plans labeled A and B, with AI tutor names hidden and A/B order randomized for each reviewer. They rated each plan from 1 to 5, gave a five-level relative preference on the 8 criteria, and explained their decision. A queue balanced coverage of pairs of AI tutors; the comparisons have one to five distinct reviewers each.

The 2,028 reviews comprise 1,136 Quantitative and 892 Verbal reviews, covering 852 distinct comparison tasks. Of the 51 reviewers, 39 reviewed Quantitative plans and 38 reviewed Verbal plans; some reviewed both. The common eight criteria yield 16,224 criterion-level evaluations: 10,140 for lesson planning and 6,084 for practice-problem creation and design. The lesson-plan comparisons cover 12 AI tutors with 13 version-specific entries; Gemini 3.6 Flash has no expert-review estimate.

For AI tutors $a$ and $b$, the Bradley--Terry model \citep{turner2012} estimates preference probability $\operatorname{logit}^{-1}(\theta_a-\theta_b)$. Ties are excluded; slight and strong preferences receive weights one and two, respectively. A pre-test is counted only if it contributes at least one preference not marked as a tie to the ranking. All 381 student pre-tests contribute to the lesson-planning ranking, but only 380 contribute to the practice ranking because reviews on one pre-test yielded only ties on the practice criteria. The plotted intervals account for repeated evaluations based on the same student pre-test. Appendix~\ref{app:reviewer_dependence} also allows for experts reviewing multiple pre-tests. Ranking scores reflect fitted expert preference.

The Quantitative test-taking-strategy criterion asked about GRE-format weaknesses in 590 reviews and broader test-taking strategy in 546. Both wordings map to the shared strategy criterion; the supplementary rubric preserves each version. The AI tutors were not all evaluated on the same student pre-tests or against the same alternatives; the ranking is based on the available comparisons.

\subsection{Repeated expert reviewers}\label{app:reviewer_dependence}
The 51 experts contributed 2,028 reviews; 39 reviewed more than one student pre-test. A sensitivity check of pooled planning, practice and eight-criterion comparisons uses uncertainty clustered by both reviewer and pre-test \citep{cameron2011multiway}. The sensitivity analysis's results are in Table~\ref{tab:reviewer_dependence}, and the analysis leaves the fitted scores, preference weights, and tie exclusions unchanged. It adds the reviewer- and pre-test-clustered covariance matrices and subtracts their intersection, using the usual finite-cluster correction for each component. Tests use a Student-$t$ reference with 50 degrees of freedom.

\begin{table}[tbp]\centering
\caption{\textbf{Expert-ranking sensitivity to repeated reviewers.} This checks the expert comparisons in Section~\ref{sec:planning_practice}. Counts show how many version-specific AI tutor entries differ significantly from Opus 5 ($p{<}.05$). The original analysis clusters uncertainty by pre-test; the sensitivity clusters by reviewer and pre-test. Fitted scores are held fixed. Ratings marked as ties are excluded.}\label{tab:reviewer_dependence}
\begin{tabular}{lrrrr}\toprule
Evaluation & Ratings & Pre-tests & Original & Sensitivity\\\midrule
Lesson planning & 9,577 & 381 & 9/12 & 8/12\\
Practice creation/design & 5,265 & 380 & 9/12 & 8/12\\
All eight shared criteria & 14,842 & 381 & 9/12 & 8/12\\\bottomrule
\end{tabular}
\end{table}

After accounting for repeated reviewers, Opus 5 no longer differs significantly from Opus 4.8 (off) in planning or the eight-criterion ranking, or from GPT-5.5 in practice design.

\subsection{Overall and criterion-specific leaderboards}\label{app:rubric_leaderboards}
\begin{figure}[!htbp]\centering
\includegraphics[width=\linewidth]{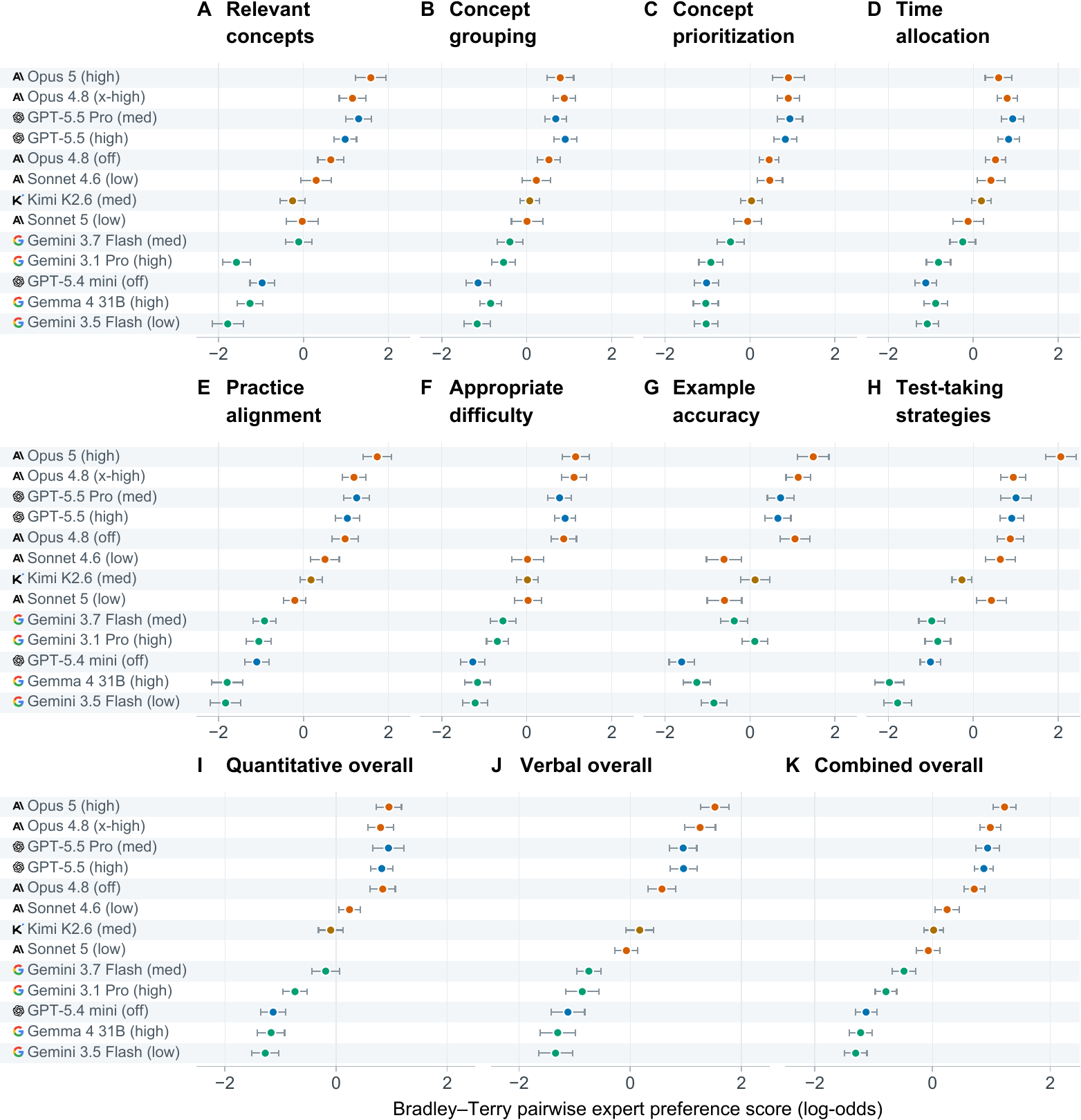}
\caption{\textbf{Expert preference by teaching criterion and overall.} A--H, the eight criteria in Table~\ref{tab:rubric}; I--K, all eight fitted jointly for Quantitative, Verbal and both sections combined, respectively. These rankings expand Figure~\figpanelref{fig:rubrics}{A, B}. Points show centered Bradley--Terry estimates; whiskers are pointwise 95\% intervals clustered by student pre-test. For each panel, we report the number of student pre-tests and expert ratings. Ratings marked as ties are excluded. A, 371 pre-tests and 1,899 ratings; B, 370 pre-tests and 1,905 ratings; C, 372 pre-tests and 1,916 ratings; D, 376 pre-tests and 1,956 ratings; E, 377 pre-tests and 1,939 ratings; F, 365 pre-tests and 1,653 ratings; G, 366 pre-tests and 1,673 ratings; and H, 378 pre-tests and 1,901 ratings. Overall fits use 246, 135 and 381 pre-tests, with 8,206, 6,636 and 14,842 ratings, respectively. All panels share the combined AI tutor order; blank cells mean not evaluated. Scores are comparable only within panels.}\label{fig:criterion_bt_grid}
\end{figure}

Figure~\ref{fig:criterion_bt_grid} expands the two leaderboards in Figure~\figpanelref{fig:rubrics}{A, B} into the eight criteria in Table~\ref{tab:rubric} and overall rankings. Panels A--H fit each criterion separately; panels I--K fit the pairwise ratings from all eight criteria together for Quantitative, Verbal and both sections combined. Each section includes 12 AI tutors and 66 pairs; the combined fit includes 13 version-specific AI tutor entries and 78 pairs. Scores are centered within each fit, so values from different panels are not directly comparable.

\clearpage
\section{Conversational pedagogy and student-reported problem quality}\label{app:conversation}
Transcripts cover all 2,139 AI sessions and 135 of the 140 human sessions. Human transcripts group consecutive lines from the same speaker into turns and include only the period from the start of tutoring to the transition to the post-test. Recordings failed for five human sessions. The conversational-pedagogy evaluation uses six rules of ``good'' tutoring synthesized from established literature in the learning and cognitive sciences \citep{wood1976, vanlehn2011, chi1994eliciting, chi2014icap, kapur2008, koedinger2007, sweller1985use, dunlosky2013improving, renkl2014toward} (see Table~\ref{tab:conversation_rules}). We then identified the prevalence of these rules using deterministic algorithms that extract wording and turn order to identify scaffolding cues, requests for explanation, interactive questions, early requests to attempt a problem, long solutions after a student reply, and reasoning checks. For each tutor, each rule's prevalence is the percentage of transcripts in which it appears. We report the mean prevalence across the six rules.

Although these rules are based on prior literature, we did not validate them as predictors of learning lift in this study.
Additionally, human sessions used spoken conversation over video calls, whereas AI sessions used typed chat. The same rules may behave differently in spoken and typed conversations; future evaluations may use a model's voice mode when available.

\begin{figure}[H]\centering
\includegraphics[width=\linewidth]{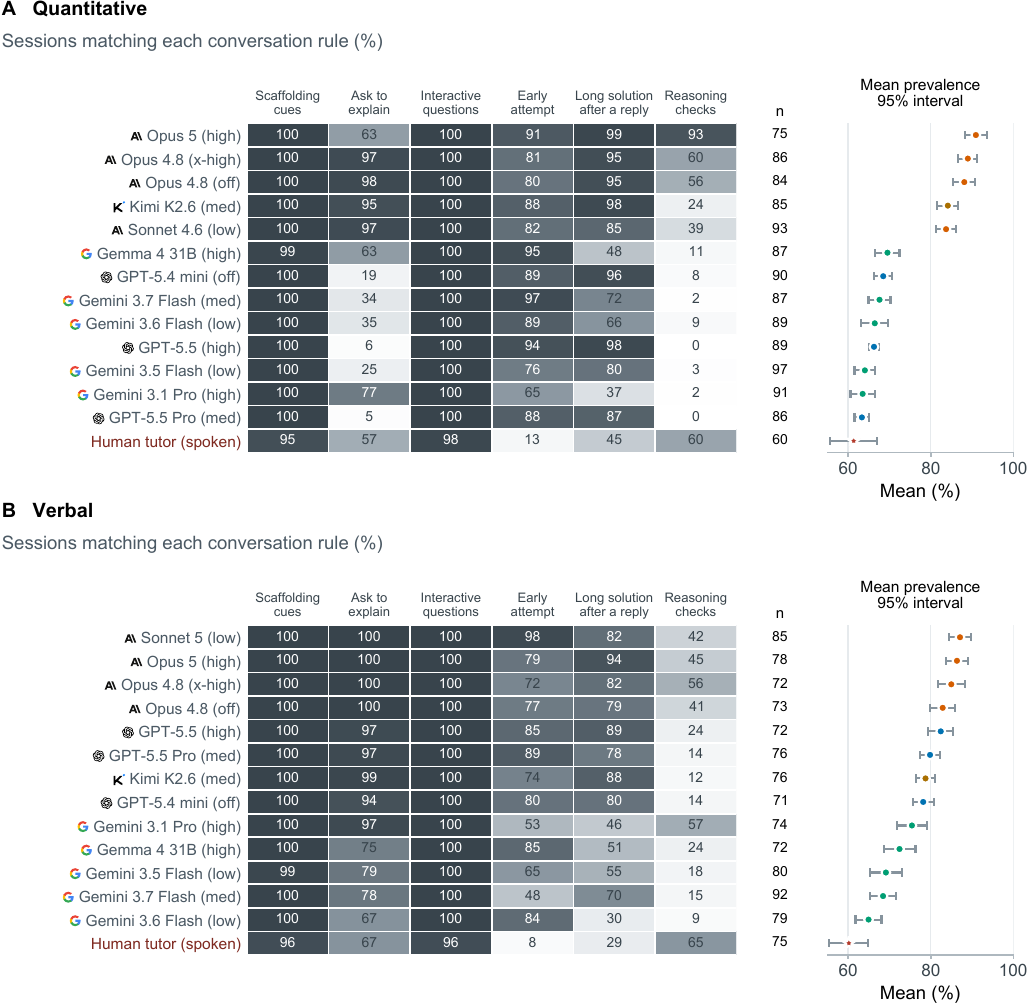}
\caption{\textbf{Conversational pedagogy by GRE section.} \textbf{A,} Quantitative. \textbf{B,} Verbal. Panels expand the Combined ranking in Figure~\figpanelref{fig:rubrics}{C} using the six rules in Table~\ref{tab:conversation_rules}. Cells give the percentage of sessions satisfying each rule; $n$ counts available transcripts. The right column shows the mean of the six indicators with 95\% Student-$t$ intervals across sessions. Human tutoring is a spoken reference and AI tutoring uses typed chat.}\label{fig:conversation_by_section}
\end{figure}

\paperfigure{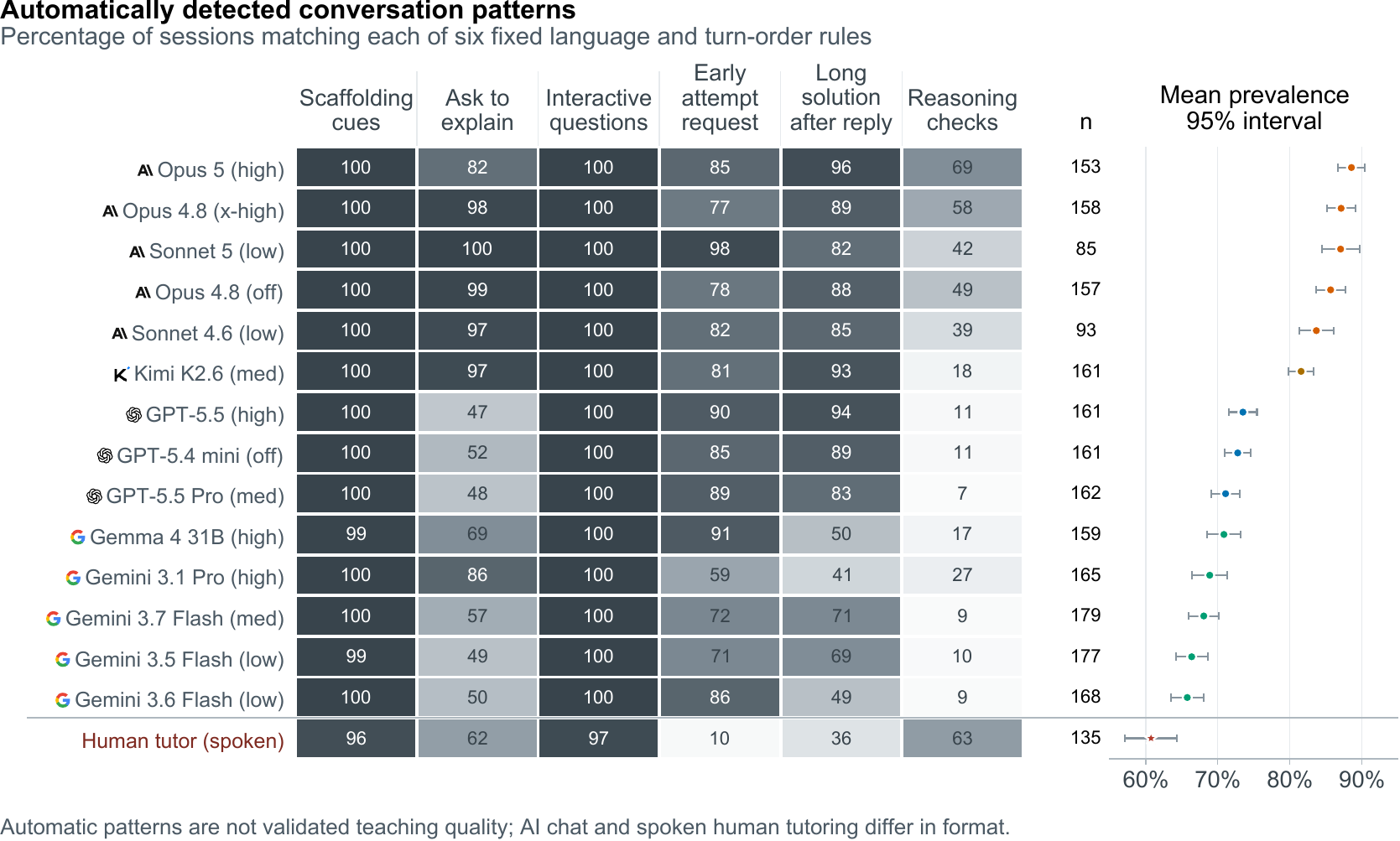}{\textbf{How tutors interact with students.} Cells show the percentage of sessions matching each rule underlying the ranking in Figure~\figpanelref{fig:rubrics}{C}. Dots show mean prevalence across the six indicators, with 95\% Student-$t$ intervals across sessions; $n$ is the transcript count. The human reference uses spoken sessions, whereas AI sessions use typed chat. See Figure~\ref{fig:conversation_by_section} for Quantitative and Verbal results and Table~\ref{tab:conversation_rules} for indicator definitions.}{fig:conversation}{0.60}

\begin{longtable}{p{0.27\linewidth}p{0.65\linewidth}}
\caption{Operational definitions for the six transcript indicators in Figures~\figpanelref{fig:rubrics}{C},~\ref{fig:conversation} and~\ref{fig:conversation_by_section}.}\label{tab:conversation_rules}\\
\toprule
Indicator & Detection rule\\\midrule\endfirsthead
\toprule Indicator & Detection rule\\\midrule\endhead
Scaffolding cues & A tutor response following a student message of at least two words contains evaluative language, or shares at least two substantive words with that message and contains a scaffolding cue.\\
Request for explanation & At least one tutor question contains a request for reasoning or explanation.\\
Interactive questions & At least four tutor messages contain a question mark or an invitation to respond.\\
Early attempt request & At least one of the first five tutor messages asks the student to try or attempt a problem.\\
Long solution after a reply & After a student message of at least two words, a tutor message of at least 140 words includes a solution or calculation cue.\\
Reasoning checks & At least two tutor questions contain a reasoning-check or strategy cue.\\\bottomrule
\end{longtable}

\subsection{Student-reported answer disputes}\label{app:student_flags}
Students could press ``I disagree with this answer --- continue'' after receiving feedback on a practice answer. For each AI tutor, the dispute rate is the fraction of distinct answered problems with at least one recorded dispute. Repeated clicks on one problem count once. The button records disagreement with the marked answer, not an independently verified error in the problem or answer key.

\begin{figure}[H]\centering
\includegraphics[width=\linewidth]{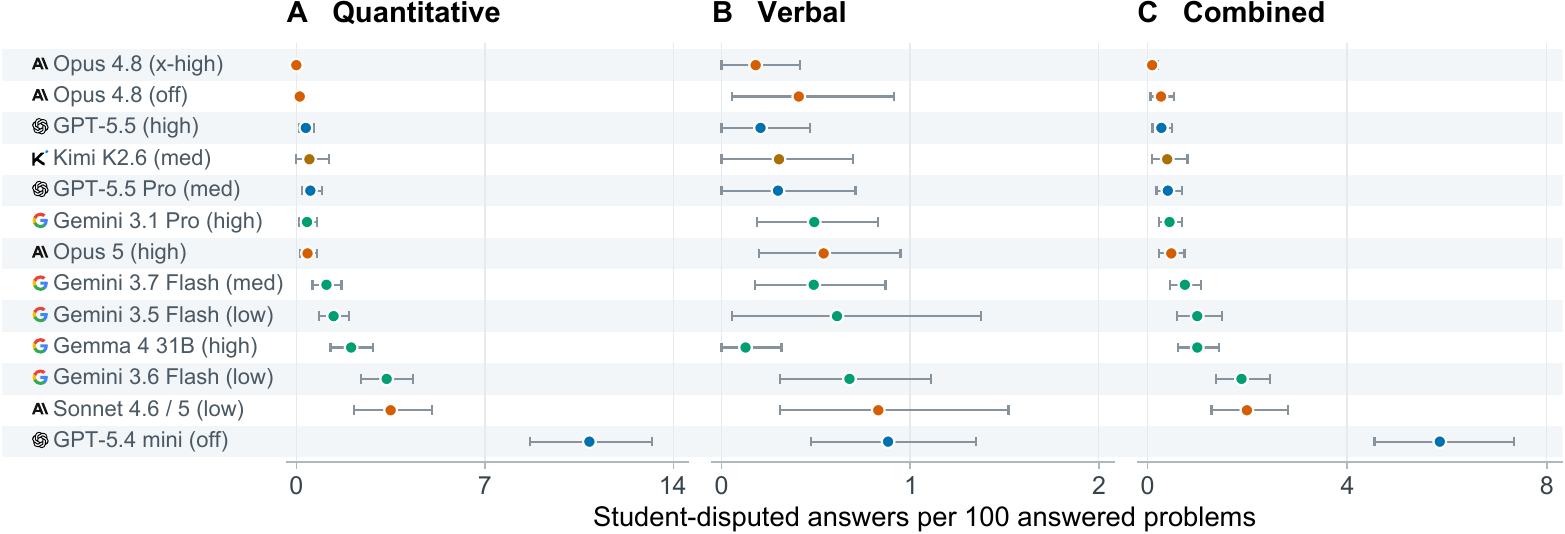}
\caption{\textbf{How often students disputed a practice answer.} A--C show flagged problems per 100 distinct answered problems in Quantitative, Verbal and Combined. Whiskers are 95\% session-bootstrap intervals across 1,139 Quantitative, 1,000 Verbal and 2,139 Combined sessions. Horizontal scales differ to show the smaller Verbal rates. Rows share the Combined ordering. The Sonnet row pools its section-specific versions in Combined. These student reports complement the expert practice rankings in Figure~\figpanelref{fig:rubrics}{B}; they record student disagreements with answers, not independently verified errors.}\label{fig:flags}
\end{figure}

Figure~\ref{fig:flags} records 523 disputed answers among 21,769 answered Quantitative problems and 120 among 23,694 Verbal problems. All observed Verbal rates were below one dispute per 100 answered problems. The 95\% percentile intervals use 4,000 session resamples, with ratios recomputed from each resample's totals. A zero observed count gives a zero-width interval but does not establish zero risk. Kruskal--Wallis tests compare per-session rates among AI tutors; the Quantitative, Verbal and Combined tests were significant ($p{<}.001$).

\subsection{Agreement with expert judgments}\label{app:flag_expert_agreement}
Higher expert preference for practice-problem creation and design (Figure~\figpanelref{fig:rubrics}{B}) was associated with fewer student disputes in Quantitative and Combined. Correlations compare AI tutor ranks for practice-problem creation and design (criteria E--G) and, separately, example accuracy (G) with the negative of each AI tutor's dispute rate. Positive correlations therefore indicate agreement between expert preferences and fewer student disputes. Each section includes 12 AI tutors with both expert ratings and student-dispute data; Combined includes 11, excluding the row that pools the two Sonnet versions. Gemini 3.6 Flash has no expert-review estimate. The section-specific expert scores are refitted from their original pairwise ratings using the same criteria as the Combined ranking.

For practice-problem creation and design, Spearman correlations were $\rho{=}0.713$ in Quantitative ($p{=}.035$), $\rho{=}0.315$ in Verbal ($p{=}.388$), and $\rho{=}0.773$ in Combined ($p{=}.033$). For example accuracy, the corresponding correlations were $0.895$ ($p{<}.001$), $0.406$ ($p{=}.388$), and $0.782$ ($p{=}.033$). These exploratory two-sided tests use 99,999 random permutations of AI tutor labels. They treat the estimated AI tutor ranks as fixed and do not account for uncertainty in those ranks. Expert review concerns examples in plans, whereas student disputes can also concern practice generated later in tutoring.

\section{Prompt-design comparisons}\label{app:extended-5}\label{app:prompt_comparisons}
The learning results in the main text and in particular in Figure~\ref{fig:learning} use the fixed main-study prompts shown in Appendix~\ref{app:prompts}. However, we did an early Quantitative pilot containing 326 sessions from 312 participants across Opus 4.8, Gemini 3.5 Flash and GPT-5.5 focusing on the prompt design.
\subsection*{Minimal and expanded prompts pilot}
\begin{figure}[H]\centering
\includegraphics[width=\linewidth]{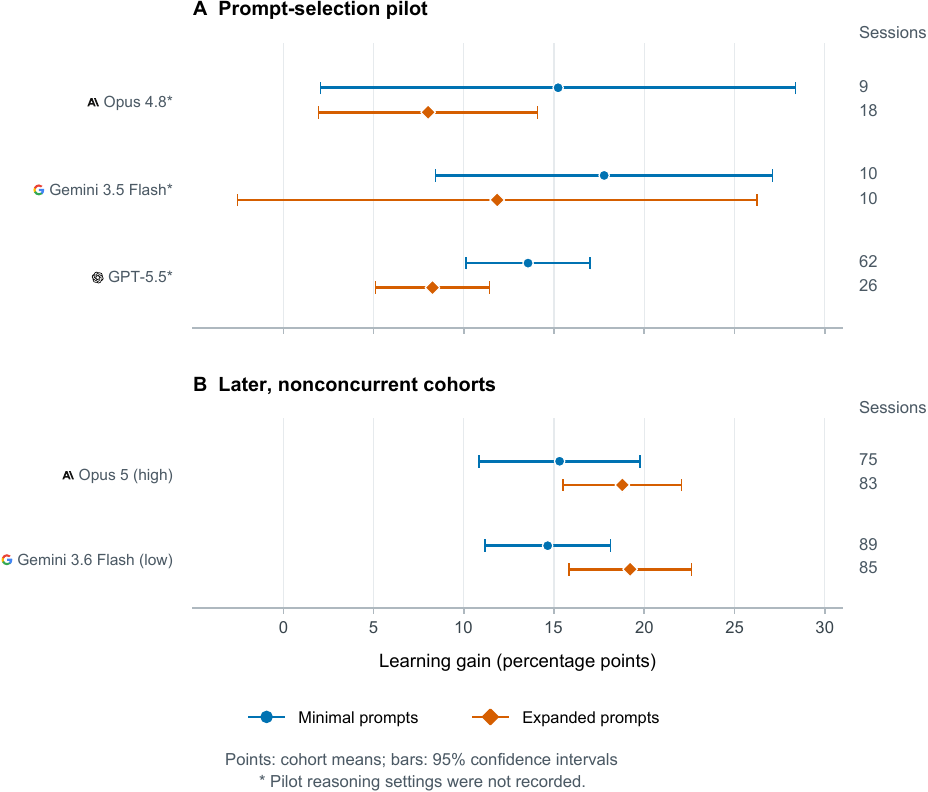}
\caption{\textbf{Learning with minimal and expanded prompts.} Supporting the prompt design in Section~\ref{sec:prompt_design}, points show mean Quantitative learning gain with separate Student-$t$ 95\% confidence intervals; counts on the right show the number of sessions. Both lesson-plan and tutor prompts are minimal for blue circles and expanded for orange diamonds. \textbf{A,} Three AI tutors in the exploratory prompt-selection pilot. This panel shows the two matched prompt settings; see Table~\ref{tab:prompt_pilot_cells} for all five settings, including mixed and intermediate prompts. \textbf{B,} Two later AI tutors, comparing supplementary cohorts that used expanded prompts with main-study cohorts that used minimal prompts. Differences are descriptive: the pilot informed prompt selection, and the later prompt conditions were not concurrently randomized.}
\label{fig:prompt_comparison}\label{fig:ext-figure-s5-quant-max-prompt}\label{fig:ext-figure-s7-prompt-pilot}
\end{figure}

 The pilot compared five combinations of minimal, intermediate and expanded lesson-plan and tutor prompts (Table~\ref{tab:prompt_pilot_cells}). When both prompts were minimal, the pooled mean gain was 14.27 percentage points, compared with 8.85 when both were expanded. Mean gains were higher with minimal prompts for each AI tutor (Figure~\figpanelref{fig:prompt_comparison}{A}). These pilot data informed the main-study prompts and the data points are excluded from the main learning cohort.

The numbers of sessions differed across prompt conditions and AI tutors, and the lesson-plan generator differed from the chat tutor in 41 sessions. An exploratory regression adjusted for pre-test score, assessment form and the AI tutor used for planning, with a full interaction between AI tutor and prompt condition and covariance clustered by first-attempt participant identity. The joint interaction was inconclusive ($\chi^2_8{=}10.00$, $p{=}.265$). None of the three tutor-specific expanded/expanded minus minimal/minimal contrasts, or their three pairwise differences, was significant.

Separate cohorts studied at different times used expanded prompts with Opus 5 and Gemini 3.6 Flash; the main-study cohorts used minimal prompts. Mean learning gains were higher with the expanded prompts for both AI tutors: 18.79 versus 15.31 percentage points for Opus 5 and 19.22 versus 14.65 for Gemini 3.6 Flash (Figure~\figpanelref{fig:prompt_comparison}{B}). After adjustment for baseline and form, the corresponding differences were 3.61 percentage points (95\% CI $[-1.66,8.89]$; $p{=}.359$) and 4.62 percentage points ($[-0.17,9.42]$; $p{=}.177$). Their interaction was also inconclusive ($p{=}.781$). The comparisons use HC3 covariance. Because the cohorts were studied at different times, these comparisons do not isolate the effect of the prompts.

\begin{table}[H]\centering
\caption{Prompt-selection evidence for the fixed-instruction design in Section~\ref{sec:prompt_design}. All five Quantitative pilot settings are shown; these sessions are separate from the main learning results in Figure~\ref{fig:learning}. Cells give mean percentage-point gain, with the number of sessions in parentheses. The two terms in each row identify the lesson-plan and tutor prompts, respectively. AI tutor columns identify the deployed chat tutor; the lesson-plan generator differed in 41 sessions.}\label{tab:prompt_pilot_cells}
\begin{tabular}{lrrrr}\toprule
Lesson plan / tutor & Opus 4.8 & Gemini 3.5 Flash & GPT-5.5 & All \\
\midrule
Minimal / Minimal & 15.23 (9) & 17.78 (10) & 13.56 (62) & 14.27 (81) \\
Minimal / Expanded & 14.81 (27) & 8.50 (27) & 12.65 (36) & 12.06 (90) \\
Intermediate / Intermediate & 11.62 (29) & 2.06 (9) & 11.40 (26) & 10.19 (64) \\
Expanded / Minimal & 8.38 (19) & 2.65 (7) & 23.23 (11) & 11.71 (37) \\
Expanded / Expanded & 8.02 (18) & 11.85 (10) & 8.26 (26) & 8.85 (54) \\
\bottomrule\end{tabular}
\end{table}

